\documentclass{article}

\usepackage{PRIMEarxiv}

\usepackage[utf8]{inputenc} 
\usepackage[T1]{fontenc}    
\usepackage{hyperref}       
\usepackage{url}            
\usepackage{booktabs}       
\usepackage{amsfonts}       
\usepackage{nicefrac}       
\usepackage{microtype}      
\usepackage{lipsum}
\usepackage{fancyhdr}       
\usepackage{graphicx}       
\usepackage{amsthm}
\usepackage{algorithm}
\usepackage{algpseudocode}
\usepackage{graphicx}
\usepackage{subcaption}
\usepackage{amsmath}
\usepackage{multirow}
\usepackage{makecell}
\graphicspath{{media/}}     

\newtheorem{theorem}{Theorem}

\newtheorem{assumption}{Assumption}

\title{Local Prediction-Powered Inference
}

\author{
  Yanwu Gu \\
  Department of Mathematics \\
  The Hong Kong University of Science and Technology \\
  Hong Kong\\
  \texttt{yanwu.gu@connect.ust.hk} \\
   \And
  Dong Xia \\
  Department of Mathematics \\
  The Hong Kong University of Science and Technology \\
  Hong Kong\\
  \texttt{madxia@ust.hk} \\
}

\begin{document}
\maketitle

\begin{abstract}
To infer a function value on a specific point $x$, it is essential to assign higher weights to the points closer to $x$, which is called local polynomial / multivariable regression. 
In many practical cases, a limited sample size may ruin this method, but such conditions can be improved by the Prediction-Powered Inference (PPI) technique.
This paper introduced a specific algorithm for local multivariable regression using PPI, which can significantly reduce the variance of estimations without enlarge the error. 
The confidence intervals, bias correction, and coverage probabilities are analyzed and proved the correctness and superiority of our algorithm.
Numerical simulation and real-data experiments are applied and show these conclusions. 
Another contribution compared to PPI is the theoretical computation efficiency and explainability by taking into account the dependency of the dependent variable.
\end{abstract}


\section{Introduction}\label{sec:intro}
In practical applications, the process of inference, particularly in describing the relationship between the dependent variable $Y$ and the independent variables $X$, remains a pivotal subject of study. 
The dependent variable $Y$ for a specific independent variable $X$ may be governed by an elusive potential function $m(X)$ that poses challenges to direct observation and precise estimation. 
At times, this function may exhibit linear characteristics in certain components or at specific points; however, it may deviate from these linear properties under different conditions. 

Subsequently, a rudimentary and straightforward regression model is inadequate for addressing the variability of parameters at local points. 
Savitsky \cite{savitzky1964smoothing} initially introduced the Savitzky-Golay filter, which is analogous to locally estimated scatterplot smoothing (LOESS), a technique later refined and expanded by Cleveland \cite{cleveland1979robust, cleveland1988locally}. 
This approach is also referred to as Locally Weighted Polynomial Regression.
By applying weights to different instances according to distance to the target point, such a model facilitates the prediction of both the value and gradient of a target at untested points, enabling the assessment of its optimality and the determination of subsequent optimization steps. 
In other words, employing a local regression model allows for the strategic planning of subsequent tests, whether to implement a temporarily optimal sample or to evaluate a more feasible treatment as guided by the model.

For instance, in the design of an industrial product with numerous features waiting to be optimized for maximal performance, a local regression model can be instrumental. 
Lin \cite{lin2017feature} proposes a model to estimate significant ship costs in the preliminary design phase. 
One of the principal cost components is influenced by the anticipated velocity and the rated power, which increase quadratically at lower velocities and cubically at higher velocities. 
Classical linear or polynomial regression models cannot estimate response variables on such shifting parameters, whereas local regression captures the information from different instances and assigns higher weights to the similar ones.
The exploration of target functions characterized by smoother curves is required by the limitation of datasets, which can be solved under local regression as well.

To describe the local regression problem, we first assume that there exists a function $m(x):\mathbb R^p\mapsto \mathbb R$ and the response variable $Y_i$ follows that
\begin{equation}\label{equ:XYrelation}
    Y_i=m(X_i) + \varepsilon_i,\ \quad \boldsymbol{\varepsilon}\sim N(0,\sigma^2I).
\end{equation}
This means that the mean value of the response variable $\mathbb E(Y_i|X=X_i)$ follows a fixed function $m(X_i)$, and the response variable $Y_i$ is associated with a noise $\varepsilon_i$, which individually follows a zero-mean Gaussian distribution.

However, as associated with previous studies such as Lu \cite{lu1996multivariate},  the expected bias and variance are influenced by the size of train set and the bandwidth parameter. 
It is intuitive that larger train set and smaller bandwidth, which will be discussed in Section \ref{sec:pre}, mean more precise information is given to the estimator and improve the performance of the regressor.
Consequently, if we want to give a precise and stable estimation of parameters, we need more data, which is hard to collect once the cost is high. 

This raises the question of whether it is feasible to expand the dataset at a reduced cost, or without direct collection or testing. 
The semi-supervised learning technique, which incorporates unlabeled data into the training process, represents one potential solution, with prediction-powered inference being a newly proposed variant.

The groundbreaking concept of Prediction-Powered Inference (PPI), ingeniously proposed by Angelopoulos et al. \cite{angelopoulos2023prediction}, advocates the use of a good predictor $F$, a product of state-of-the-art machine learning algorithms, to bestow a prediction on an unlabeled dataset $\mathcal U$. 
This predictor allows treating the unlabeled dataset as pseudo-labeled one, estimating parameters and their confidence intervals. 

PPI uses the predictor $F$ to expand the richness of the datasets, which will decrease the variance taken by the number of sample sizes from $O(n^{-1})$ to $O(N^{-1})$ where $n, N$ is the sample size of the labeled dataset and the unlabeled dataset, respectively.
Meanwhile, the volatility led by the predictor, although with coefficient $O(n^{-1})$, is much lower than that of the original estimator, because the predictions of the good predictor have more stable and smaller errors.
At last, a rectifier will correct the bias lead by the predictor $F$.

Take the force estimation of ship design as an example again: The classical ship dataset includes a few or even no ships with parameters similar to our designs, which will make our local regression model invalid if the dimension is relatively high.
With PPI, we only need to design some data that are close to the target point in terms of parameters, or even close to several parameters, so that the effect of local regression can be improved compared with that of not being used, as long as the effect of the predictor $F$ is good enough.

Throughout the estimation process, despite our presumption of the efficacy of the predictor $F$, the use of the prediction of an unlabeled dataset can lead to estimation bias.
The algorithm concurrently employs a rectifier $\Delta$ to counterbalance the impact of the estimation bias caused by $F$.

We employ prediction-powered inference within the context of local multivariable regression and conduct inference on two real-world datasets: water quality and air quality. 
For each target point, we analyze the standard error of the estimation error and the mean squared error / accuracy against the ground truth value, thereby contrasting conventional local multivariable inference with local prediction-powered inference. 
It is important to note that the terms 'decrease' and 'increase' refer to the changes induced by prediction-powered inference on local inference.
\begin{figure}[htb]
\centering
\subfloat[Water Quality Inference: Decreasing variance occurs in 84.9\% of cases, with 55.6\% improving the accuracy (Red) and 29.3\% decreasing (Green), while increasing variance occurs in 15.1\%, with 5.1\% improving accuracy (Yellow) and 10.1\% increasing (Blue).]{\includegraphics[width=0.48\linewidth]{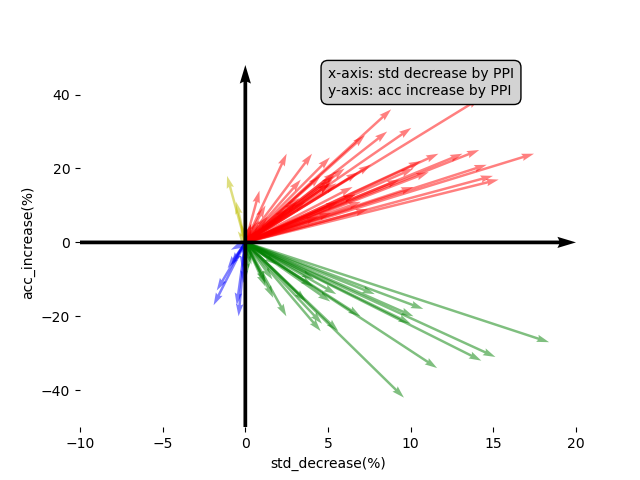}}
\hfill
\subfloat[Air Quality Inference: Decreasing variance occurs in 69.5\% of cases, with 38.2\% reducing MSE (Red) and 31.3\% increasing (Green), while increasing variance occurs in 30.5\%, with 15.4\% reducing MSE (Yellow) and 15.1\% increasing (Blue).]{\includegraphics[width=0.48\linewidth]{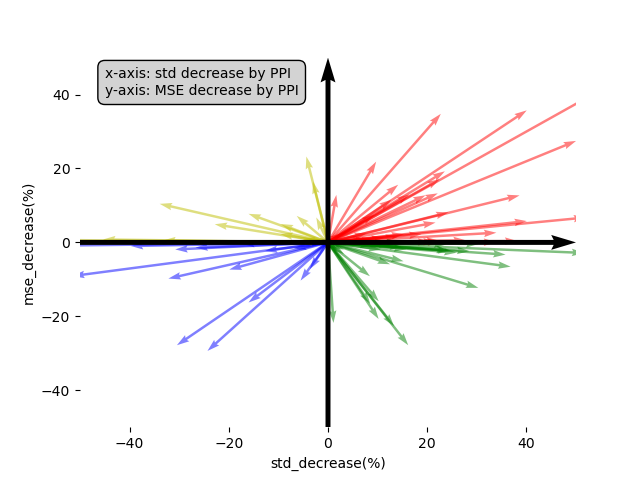}\label{fig:air_sub_pie}}
\caption{PPI Improvement Outline}
\end{figure}
From the figure of arrows, it can be inferred that the estimation for 70 percent of the instances shows an improvement in stability, indicated by a decrease in the standard deviation, without compromising accuracy, as the proportion of instances with an increase in mean squared error does not exceed those with a decrease.
Further discussion of the air quality dataset is conducted in Section \ref{subsec:air_quality}.

There also exist several problems with PPI algorithms.
First, the common algorithm paradigm for the convex estimation problem, Algorithm 5 of Angelopoulos et al.\cite{angelopoulos2023prediction}, computed the gradient estimator $g_\theta$ and the rectifier $\Delta$ using every sample but once a time, which is unsolvable or heavily biased for many problems, including the local polynomial regression problem.
Second, the algorithm only considered separate components of the parameters. This implies that the algorithm cannot use the information of other components, even if some of them have very strong confidence, which can help to estimate others.
Third, PPI mainly considers the condition of global estimation rather than local properties.
Lastly, but most importantly, there is no specific criterion to evaluate whether a predictor is good or not.

The paper is organized as follows. 
Section \ref{sec:pre} gives some preliminaries and literature on the local polynomial regression problem and prediction-powered inference. 
Section \ref{sec:theo} gives the main algorithm for local prediction-powered inference and asymptotic analysis, with the confidence region, bias correction, and coverage probability mentioned. 
Section \ref{sec:exp} uses several simulation experiments and real-world datasets to prove the theorems proposed before and compares them with the traditional local polynomial / multivariable regression methods.
The main conclusions and contributions of this paper are contained in Section \ref{sec:con}.

\section{Related Works}\label{sec:pre}

\subsection{Multivariable Local Linear Regression}
Assume that the second derivative of $m(x):\mathbb R^{p}\mapsto \mathbb R$ exists, then given a feasible feature $x$ which we want to estimate, we can write the Taylor expansion of $m(x)$ as 
\begin{equation}
\begin{split}
    m(X_i) = m(x) + \nabla m(x)^T(X_i-x) + (X_i-x)^T\nabla^2 m(\xi_i)(X_i-x),\\
\end{split}    
\end{equation}
where $\xi_i = x+t(X_i-x),t\in(0,1)$ and $X_i$ is a point in the neighborhood region of $x$. 
If $X_i$ is close enough to $x$, then the second-order term can be omitted, that is, $m(X_i) \approx  m(x)+\nabla m(x)^T(X_i-x)$. 
By replacing $m(X_i)$ with the label $Y_i$, we have $\varepsilon_i = Y_i - m(x) - \nabla m(x)^T(X_i-x)$. 
Since the expected value of the noise $\varepsilon_i$ is zero, we can use our labeled dataset $\mathcal L=\{(X_i, Y_i),i\in[n]\}$ to estimate $m(x)$, as well as $\nabla m(x)$, which can be used to analyze the influence and importance of each component around the neighborhood region of $x$. 
The methodology is to solve the following optimization problem:
\begin{equation}\label{equ:optprob_linear}
    \arg\max_{a\in\mathbb R,b\in\mathbb R^p}\sum_{i=1}^n \left(Y_i - a - b^T (X_i-x)\right)^2,
\end{equation}
which reduces linear regression. 
In the optimization problem Equation \eqref{equ:optprob_linear}, $a$ is an estimator of $m(x)$, and $b$ is an estimator of $\nabla m(x)$.
According to our assumption, only when the samples are in sufficient proximity to the target $x$ can the second-order term be ignored. 
The estimations of $m(x)$ and $\nabla m(x)$ improve with the closeness of $X_i$ and $x$. 
Thus, we should give higher weights to the closer samples, using weight functions $K(\cdot)$ and bandwidth $h$.

To reach the requirements, the weight function should be non-negative, continuous, symmetric, and decreasing on $[0,\infty)$ supported by 
Loader \cite{loader2006local}.
The non-negativity characteristic guarantees that every individual sample will not detrimentally influence the estimation. 
In the event that the weight descends into a negative value, the consequential substantial estimation error on such samples will mitigate the loss, potentially resulting in an unbounded loss and solutions that are not within acceptable parameters. 
The computation and analysis of estimation can be simplified by the presence of continuity and symmetries. 
The monotonic nature of the function ensures that samples closer to $x$ have a greater contribution, while those further away contribute less, potentially even nothing.

To sum up, the optimization problem of local multivariable regression problem under $n$ labeled data instances can be described as 
\begin{equation}
    \left(\begin{matrix}
        \widehat{m(x)}\\\widehat{\nabla m(x)}
    \end{matrix}\right) = \arg\min_{\theta\in\mathbb{R}^{p+1}} \sum_{i=1}^n\left(Y_i- (X_i-x)^{+T}\left(\begin{matrix}
        \widehat{m(x)}\\\widehat{\nabla m(x)}
    \end{matrix}\right)\right)^2K\left(\frac{X_i-x}{h}\right),
    \label{equ:abexp}
\end{equation}
where $(X_i-x)^+ = (1,\ (X_i-x)^T)^T$ is augmented feature and $h$ represents the bandwidth parameter, a critical determinant in controlling the degree of smoothness, as proposed by Fan et al. \cite{fan1996study}.
By the solution of this optimization problem, we have estimators of $m(x)$ and $\nabla m(x)$:
\begin{equation*}\label{mestimate}
    \widehat{m(x)} = \theta_1,\quad \widehat{\nabla m(x)} = \theta_{2:p+1}.
\end{equation*}
And we also define the ground-truth target as $\theta^*$, that is
\begin{equation}
    \label{target}
    \theta^* = \left(\begin{matrix} m(x)&\nabla m(x)
    \end{matrix}\right)^T.
\end{equation}
Given a particular labeled dataset denoted as $\mathcal{L}$, the anticipated parameter formulation can be reconstituted as Equation \eqref{equ:ablabeled}, located in Section \ref{subset:assump}. 
Through our analysis, although estimation under a labeled dataset asymptotically approaches the expectation solution and converges to the target value in probability, the covariance of the estimator will increase if the sample size of $\mathcal{L}$ is small. 
In other words, the limited number of samples will induce high volatility in parameter estimation, a phenomenon prevalent in contemporary AI for scientific problems \cite{gunnarsson2024prediction, cai2021unified}. 
Following the introduction of local regression, a wide variety of acclaimed kernel, spline, and orthogonal series methodologies have been developed for the estimation of $m(\cdot)$, including prominent examples such as Nadaraya-Watson\cite{nadaraya1964estimating, watson1964smooth} and Gasser-Müller\cite{gasser1979kernel}. 

An assumption relevant to the properties of target function and point has been articulated by Fan et al.\cite{fan1993local}.

\begin{assumption}\label{assump:fanlocal}
\vspace{0.5em} 
    (i) The regression function $m(\cdot)$ has a bounded second derivative.
    
    (ii) The marginal density $f(\cdot)$ of $\mathcal X$ satisfies $|f(x)-f(y)|\leq c\|x-y\|^{\alpha}$, for $0<\alpha<1$, and $f(x_0)>0$ where $x_0$ is the point of interest. There is an open neighborhood $U$ of $x_0$ such that $m\in C^3(U), f\in C^1(U)$.

    (iii) The conditional variance $\sigma^2(x)=\text{Var}(Y|X=x)$ is bounded and continuous. This condition holds because of the constant variance.
\end{assumption}

Based on Theorem 1 and 2 of Fan et al.\cite{fan1993local}, if $h=dn^{-\beta}$, $0<\beta <1$, then the estimator Equation \ref{mestimate} satisfies 
\begin{equation*}
    \mathbb E(\widehat{m(x)}-m(x))^2 = O(h^4+(nh)^{-1}).
\end{equation*}

The topic of how to choose a proper kernel function $K(\cdot)$ to reach minimax efficiency has been studied for long.
Gasser \cite{gasser1985kernels} investigated the choice of kernels for the nonparametric estimation of regression functions and of their derivatives, which is then widely used in local regression methods and in estimating the probability density function and spectral densities.  
Then Fan \cite{fan1993local} proved that the univariate local linear regression estimator exhibits commendable sampling properties and superior minimax efficiency both in rates and constant factors, epitomizing the optimal linear smoother and attaining the asymptotic linear minimax risk.
Subsequently, Fan \cite{fan1997local} extended this framework to encompass multivariable local linear regression and polynomial linear regression estimators. This work introduced an optimal kernel for local multivariable regression, thus elucidating the existence of a universally optimal weighting scheme.

For consistency in the estimation of the gradient, Lu \cite{lu1996multivariate} suggested the following assumption of the kernel function $K(\cdot)$:

\begin{assumption}\label{asump:multi}
    The kernel $K(\cdot)$ is a spherically symmetric density function, that is, there exists a univariate function $k(\cdot)$ such that $K(\mathbf{x})=k(\|\mathbf x\|)$ for all $\mathbf x\in\mathbb R^p$. Furthermore, we assume that the kernel $K(\cdot)$ has an eight-order marginal moment, that is, $\int u_1^8 K(u_1,\dots, u_p)du_1\dots du_p<\infty$. Consequently, the odd-ordered moments of $K$ and $K^2$, when they exist, are zero; i.e., for $l=1, 2$
\begin{equation*}
    \int u_1^{i_1}\dots u_p^{i_p} K^l(u)du = 0,\ \text{ if }\ \sum_{i=1}^pi_p\ \text{ is odd.}
\end{equation*}

\end{assumption}

Another critical issue in nonparametric smoothing techniques is the selection of the bandwidth or smoothing parameter. 
An excessively large bandwidth results in an insufficient number of effective training samples, whereas an overly small bandwidth assigns equal weight to samples of varying significance, thereby undermining the essence of local regression.
Fan and Gijbels \cite{fan1995data, fan1996study} analyzed the empirical performance of proposed fully-automatic bandwidth selection procedure and derived the asymptotic result by balancing the bias and variance, which obtain the optimal variable bandwidth. 
Ruppert \cite{ruppert1995effective} implemented the principles of plug-in bandwidth selection to formulate methodologies to determine the smoothing parameter of local linear least squares kernel estimators. These methodologies are pertinent to odd-degree local multivariable fits and possess the potential for extension to various other contexts, including derivative estimation and multiple nonparametric regression.

\subsection{Prediction-Powered Inference}

Recall the procedure of prediction-powered inference, the nature of this method is to use the unlabeled dataset $\mathcal U$ to do the inference and then fix the bias brought by the unlabeled data and predictor $F$ using the rectifier estimated on the labeled dataset $\mathcal L$. 
The idea of using unlabeled data to expend the information is actually semi-supervised learning. 
However, we do not use labeled dataset directly, but fix the bias post-prediction, that is, de-bias on the inference result of unsupervised learning.

The methodology for rectifying statistical inference utilizing outcomes predicted by an arbitrarily selected machine learning algorithm was initially introduced by Wang et al. \cite{wang2020methods}.
They divide the dataset into training, testing, and validation sets. 
The model is trained on the training subset, and the relationship between the observed and predicted outcomes was estimated on the testing subset. 
This estimated relationship is subsequently employed to adjust inference in the validation subset.

Despite the flexibility of Wang's methodology, which employs post-prediction-adjusted point and interval estimates and is applicable to both continuous and categorical outcome data, thereby mitigating the impact of variability and bias more effectively than intuitive approaches, this technique is contingent upon the relationship between observed and predicted outcomes. 
When the model fails to accurately capture this relationship, bias correction is insufficient to yield valid inferences. 
In essence, even with a robust predictor, the relationship between observed data and prediction outcomes may remain elusive. 
Simplistic assumptions are inadequate to resolve this issue.

Fortunately, Angelopoulos et al.  \cite{angelopoulos2023prediction} find a way, i.e. Prediction-Powered Inference, to avoid the estimation between the observed and prediction data. In their approach, they neglect the training process and assume the existence of a good predictor $F$. 
Rather than providing predictions for labeled data and examining the relationship between true labels and predicted labels, prediction-powered inference eschews the use of labeled data and instead focuses on predicting unlabeled data. 
In one respect, acquiring unlabeled data is more feasible than annotating various data points, particularly for statisticians who concentrate on specialized issues in other fields. 
On the other hand, reallocating data from the validation set to the test set (while maintaining an empty training set) augments the scope of inference, mitigates bias and variance, and thereby renders the inference more robust.

In the correction procedure, Wang's method employs bootstrap techniques to estimate parameter values and their standard errors, ultimately selecting the medians of the bootstrap outcomes as the final estimates. 
While this method achieves de-biased results, it also leads to information loss. 
In contrast, prediction-powered inference (PPI) constructs a rectifier by leveraging the discrepancy between estimations of true labels and predicted labels, relative to a fixed constant. 
In their study, inference targets for mean estimation, quantile estimation, logistic regression, and linear regression are equivalently transformed into convex optimization problems, ensuring that the gradient of the optimal solution remains zero. 
By incorporating various labeled and unlabeled data together with their predictions into the gradient expressions of individual samples, PPI identifies parameters that are feasible within the confidence level $\alpha$ as the confidence set, thereby completing the inference process. 
This approach ensures that each sample is equally weighted, maximizing the utilization of available data.

Prediction-powered inference is actually a technique of semi-supervised learning. 
Implementing semi-supervised learning into inference can be traced back to the 20th century \cite{robins1994estimation, ruppert1995effective}. They proposed the estimation of the regression coefficient and multivariate models under the assumption of missing data, that is, unlabeled instances, respectively, and then prove the efficiency of such semi-supervised techniques.
More statistical inference and estimation tasks include mean \cite{Zhang2016SemisupervisedIG}, quantile estimation \cite{chakrabortty2022semi},  linear regression of general settings \cite{azriel2022semi, chakrabortty2018efficient} and high-dimensional condition \cite{zhang2022high}, and M estimation \cite{song2024general} have been proposed in recent years.

However, the general algorithm, Algorithm 5 Prediction-powered convex estimation of \cite{angelopoulos2023prediction}, faces some challenges as well. 

Firstly, the constant PPI use for inference is the zero gradient under the acceptable confidence sets. 
Such a set is not cognitive, or, does not have an explicit close form solution. 
As a result, the confidence set can only be formulated as
$$
\mathcal C_\alpha^{\text{PP}} = \left\{\theta \in \Theta_{\text{grid}} : |\hat g^f_{\theta,j} + \hat\Delta_{\theta,j}| \leq w_\alpha(\theta)\right\},
$$
where $\Theta_{\text{grid}}$ is the fine grid where we do the parameter search, $\hat g^f_{\theta,j}$ and $\hat\Delta_{\theta,j}$ are corresponding components of gradient and rectifier, which are supposed to be zero. 
And $w_\alpha(\theta)$ is the test statistic with respect to the estimated error.
It is apparent that the construction of such a confidence set requires a lot of computing resource, and the computation of individual components does not make full use of the information.

To conquer this problem, PPI++ \cite{angelopoulos2023ppi++} proposes new approaches to tackle the inference problems of generalized linear model and M-estimation problems.
They replace the variance estimated by the bootstrap method with the multiplication of inverse of Hessian matrix and covariance of the gradient. 
Such approaches consequently estimate the width of confidence intervals under increased usage of information.

Secondly, there is no definitive criterion to assess the adequacy of a predictor $F$ for implementation in PPI. 
In Wang \cite{wang2020methods} and Angelopoulos \cite{angelopoulos2023prediction}, no explicit criterion is provided to select a machine learning algorithm from a pool. 
However, the efficacy of both methods is contingent upon the performance of the machine learning algorithm on the dataset of interest to statisticians, specifically in terms of accuracy and consistency. 
Regarding accuracy, it is evident that pseudo-labels significantly deviating from the true labels will result in erroneous estimations, even with bias correction methods or rectifiers. 
Furthermore, if the predictor lacks consistency, that is, the variances of errors across different estimation targets vary substantially, these points may exhibit erratic fluctuations with incorrect predictions, ultimately leading to inaccurate estimations.

Finally, post-prediction data-driven methodologies are predicated on the assumption of dataset consistency to mitigate bias. The identical data-generating process substantiates the uniform expectation of variables, thereby ensuring that the rectification and correction methods can bridge the gap before and after the application of the prediction label. In cases where datasets are inconsistent, Wang \cite{wang2020methods} recommends the following approach:
\begin{enumerate}
\item Implement data normalization using techniques such as surrogate variable analysis \cite{leek2007capturing},
\item Eliminate unwanted variation \cite{risso2014normalization},
\item Address Batch Effect in linear models for micro-array data to rectify latent confounders in the testing or validation sets.\cite{smyth2005limma}
\end{enumerate}

\section{Theory}\label{sec:theo}
\subsection{Preliminaries}\label{subset:assump}

Suppose that we have a labeled dataset $\mathcal{L}=\{(X_i,Y_i),i\in[n]\}$, an unlabeled dataset $\mathcal{U}=\{(\widetilde X_i,\widetilde Y_i),i\in[N]\}$, where $\widetilde Y_i$ are unknown and $N\gg n$. 
The features $X_i$ and $\widetilde X_i$ are independently and identically distributed (i.i.d) from a distribution $\mathcal{X}$. 
The relationship between response values $Y$ and $\widetilde Y$ and the features $X$ and $\widetilde X$ satisfies Equation $\eqref{equ:XYrelation}$, and the potential function $m(x)$, marginal density $f(\cdot)$, and weight function $K(\cdot)$ all satisfy the condition of Assumption \ref{assump:fanlocal} and Assumption \ref{asump:multi}.

The condition of i.i.d. pertaining to $X_i$ and $\widetilde X_i$ inherently suggests the i.i.d. of $Y_i$ and $\widetilde Y_i$. 
The uniformity of this distribution guarantees that any estimation of a given parameter, provided they are predicated on $X_i$ ($\widetilde X_i$) and $Y_i$ ($\widetilde Y_i$), will possess a consistent expectation. 
Consequently, the terms of estimation predicated on $X_i$ can be supplanted by the equivalent term predicated on $\widetilde X_i$, thereby potentially reducing the variance. 
Furthermore, the terms predicated on $\widetilde Y_i$, of which we are unaware, can be approximated by $Y_i$.

Another assumption is that the good predictor $F$ required by prediction-powered inference is given rather than trained by the observed dataset $\mathcal{L}$ and $\mathcal{U}$.
In other words, the predictor $F$ is independent of $\mathcal{L}$ and $\mathcal{U}$.
In traditional machine learning settings, algorithms make predictions based on the training set $\mathcal L$, which also learns the noise value of $\mathcal L$. 
Consequently, the deliberately introduced information of $\mathcal L$ would lead to biased estimates of the target parameter and the rectifier $\Delta$.

In asserting the preeminence of $F$, we postulate that the anticipated discrepancy of the prediction, in relation to the authentic function $m(x)$, is significantly smaller compared to the actual value of the function.
\begin{assumption}\label{assump:n_ratio}
    The residual of the predictor $F$ with respect to $m(x)$ satisfies
    \begin{equation}
        \mathbb E[F(X_i) - m(X_i)]^2 \ll \mathbb E[m(X_i)]^2
    \end{equation}
\end{assumption}

\subsection{Conventional Estimation}\label{subsec:int_est}

To solve Equation \eqref{equ:abexp} under the labeled dataset $\mathcal L$, we take the derivative of the loss function and obtain the solution as

\begin{equation}
    \left(\begin{matrix}\widehat{m(x)}^{\text{con}}&\widehat{\nabla m(x)}^{\text{con}}\end{matrix}\right)^T = \widehat\theta^{\text{con}}
    = \arg\min_{\theta\in\mathbb R^{p+1}}\|\mathbf{W}^{1/2}(\mathbf{Y}-\mathbf{X}^T\theta)\|_2^2
    \label{equ:ablabeled}
\end{equation}
where
\begin{equation*}
    \mathbf{W} = \text{diag}\left(K\left(\frac{X_1-x}{h}\right),\dots, K\left(\frac{X_n-x}{h}\right)\right),\quad \mathbf{Y} = \left(\begin{matrix} Y_1\\\vdots\\ Y_n \end{matrix}\right),\quad \mathbf{X} = 
    \left(\begin{matrix}1 &\cdots &1\\ (X_1-x)_1 & \cdots & (X_n-x)_1\\ \vdots  &\ddots & \vdots \\ (X_1-x)_p & \cdots & (X_n-x)_p\\\end{matrix}\right)
\end{equation*}

and the superscript \textit{con} stands for conventional. Letting the derivative of the loss function be zero, we have the explicit expression in the following equation.
\begin{equation}
    \widehat\theta^{\text{con}}
    = (\mathbf{X}\mathbf{W}\mathbf{X}^T)^{-1}\mathbf{X}\mathbf{W}\mathbf{Y}
    \label{equ:ablabeled_solved}
\end{equation}

Lu \cite{lu1996multivariate} proposed the following theorem to estimate the expected error:
\begin{theorem}\label{thm:oneorder-error}
Under Assumption \ref{asump:multi}, for $h=n^{-\beta}, 0<\beta<p^{-1}$ as $n\to\infty$, the conditional bias of local linear regression and derivative given by the solution of Equation \eqref{equ:abexp} have the asymptotic expansions as
\begin{equation}
    \mathbb E\left\{\left(\begin{matrix} \widehat{m(x)}^{\text{con}} \\ \widehat{\nabla m(x)}^{\text{con}} \end{matrix} \right)-\left( \begin{matrix} {m(x)} \\ {\nabla m(x)} \end{matrix}\right)\right\} =B(x,h)+O(h^4)+O(n^{-1/2}h^{2-p/2})
\end{equation}
where 
\begin{equation*}
    \begin{split}
        B(x,h) &= h^2\left(\begin{matrix}
            \frac12f(x)\mu_2\text{Tr}(\nabla^2m(x))\\
            \frac{1}{2\mu_2f(x)}b_1(m) +\frac{1}{3!\mu_2}b(m)
        \end{matrix}\right),\\
        b(m) &= \int u D_m^3(x,u) K(u) du,\\
        b_1(m) &= \int u [u^T \nabla^2 m(x) u][\nabla f^T(x)u]K(u)du-\mu_2^2  \nabla f(x)\text{Tr}(\nabla^2 m(x)),\\
        \mu_l &= \int u_1^l K(u) du,\\
        D_g^k(x,u) &= \sum_{i_1+\dots+ i_p=k}\frac{k!}{i_1!\dots i_p!}\frac{\partial^k g(x)}{\partial x_1^{i_1}\dots\partial x_p^{i_p}}u_1^{i_1}\dots u_p^{i_p}.
    \end{split}
\end{equation*}
The covariance of estimation can be described as 
\begin{equation}
    \text{Cov}\left(\left(\begin{matrix}        \widehat{m(x)}^{\text{con}}&\widehat{\nabla m(x)}^{\text{con}}    \end{matrix}\right)^T\bigg|X_1,\dots, X_n\right) = \frac{\sigma^2}{nh^pf(x)}\left\{\left(\begin{matrix}J_0&\\ &\frac{J_2}{\mu_2^2h^2}I_p\end{matrix}\right)+O(h^{2}) + O(n^{-1/2}h^{-p/2})\right\}
\end{equation}
where $J_i=\int u_1^iK^2(u)du$.
\end{theorem}
To simplify, the expected bias is $O(h^{2}+n^{-1/2}h^{(2-p/2)})$, which is asymptotically equivalent to $0$ as $n\to\infty$. 
The proof of the initial paper is left out, and we include it in Appendix \ref{appsubsec:thm_oneorder_error}.

If we only have the labeled data $\mathcal{L}$, then the solution of $\widehat{\theta}^{\text{con}}$ in Equation \eqref{equ:ablabeled} is the best estimation of $a^*,b^*$ in Equation \eqref{equ:abexp}, that is, $\widehat{m(x)}$ and $\widehat{\nabla m(x)}$. However, the cost of constructing $\mathcal{L}$ may be too high to afford, and if we have a good predictor $F$, the unlabeled dataset $\mathcal{U}$ can also be used to construct the confidence interval of $\theta^*$.

\subsection{Local Prediction-Powered Inference Estimator}

In the study by Angelopoulos \cite{angelopoulos2023prediction}, PPI employs each individual sample within the set $\mathcal U$ to formulate an aggregate approximation of the parameter $\theta^*$. This methodology is deemed unsuitable for the local prediction-powered inference problem, because $\mathbf X\mathbf W\mathbf X$ becomes singular if the number of samples in $\mathbf{X}$ is less than the number of dimensions, that is, $n<p$. 

However, it is possible to initially approximate $\theta^*$ in the context of the feature of $\mathcal U$, and consider the forecast of $F(\widetilde X_i)$ as the response variable $\widetilde{Y}_i$. 
We then rectify the bias introduced by this approximation under the $\mathcal L$ and its corresponding pseudo-label produced by $F$, which is inspired by the rectifier $\Delta$ in PPI.

By implementing Equation \eqref{equ:ablabeled_solved} to $\mathcal{U}$, we have 
\begin{equation}
    \begin{split}
        \left(\begin{matrix}\widehat{m(x)}^{\text{con}}_{(N)} & \widehat{\nabla m(x)}^{\text{con}}_{(N)} \end{matrix}\right)^T
        =\widehat{\theta}^{\text{con}}_{(N)}=&\ \arg\min_{\theta\in\mathbb R^{p+1}}\|\mathbf{\widetilde W}^{1/2}(\mathbf{\widetilde Y}-\mathbf{\widetilde X}^T\theta)\|_2^2\\
        =&\ (\mathbf{\widetilde X}\mathbf{\widetilde W}\mathbf{\widetilde X}^T)^{-1}\mathbf{\widetilde X}\mathbf{\widetilde W}\mathbf{\widetilde Y}
    \end{split}
    \label{equ:abunlabeled_solved}
\end{equation}
where $\mathbf{\widetilde W}$, $\mathbf{\widetilde Y}$, $\mathbf{\widetilde X}$ and corresponding $\mathbf{W}$, $\mathbf{ Y}$, $\mathbf{ X}$ of labeled dataset $\mathcal {L}$ on unlabeled dataset $\mathcal {U}$.

Although the conventional estimation under $\mathcal{U}$ reduces the variance by increasing the sample size, it contains the information of $\widetilde Y$, which is unknown under our assumption. 
Thus, we should use $F(\widetilde X_i)$ to replace $\widetilde Y_i$, and use rectifier to balance the bias taken by this replacement.
We denote the bias of estimation taken by the good predictor $f$ as the rectifier
\begin{equation}
    \widehat\Delta_{(n)} = (\mathbf X\mathbf W\mathbf X^T)^{-1}\mathbf X\mathbf W \left(\begin{matrix} F(X_1)-Y_1\\\vdots\\F(X_n)-Y_n \end{matrix}\right)
\end{equation}
and use it to substitute the bias taken by the unknown $\mathbf{\widetilde Y}$, that is, $\widehat\Delta_{(N)}$:
\begin{equation}    
    \begin{split}
        &\widehat{\theta}^{\text{con}}_{(N)}=(\mathbf{\widetilde X}\mathbf{\widetilde W}\mathbf{\widetilde X}^T)^{-1}\mathbf{\widetilde X}\mathbf{\widetilde W}\mathbf{\widetilde Y}_F-\widehat \Delta_{(N)} \\
        \approx  &\widehat{\theta}^{\text{PP}} =(\mathbf{\widetilde X}\mathbf{\widetilde W}\mathbf{\widetilde X}^T)^{-1}\mathbf{\widetilde X}\mathbf{\widetilde W}\mathbf{\widetilde Y}_F-\widehat\Delta_{(n)}
    \end{split}
\end{equation}
where $\mathbf{\widetilde Y}_F = \left(F(\widetilde X_1),F(\widetilde X_2),\cdots,F(\widetilde X_N) \right)^T$.
As we assumed, the identical distribution guarantees the same expectation.

\begin{theorem}\label{thm:estimation}
    Under assumption of Theorem \ref{thm:oneorder-error}, Assumption \ref{assump:n_ratio}, and let $N \gg n$, we have 
\begin{equation}
    \begin{split}
        \mathbb E\left(\left(\begin{matrix} \widehat{m(x)}^{\text{PP}}\\\widehat{\nabla m(x)}^{\text{PP}}  \end{matrix}\right)-\left(\begin{matrix} {m(x)} \\ \nabla m(x)   \end{matrix}\right)\bigg|\mathcal L,\mathcal U\right) &= B(x,h) + \left(\begin{matrix}1\\&h^{-1}I_p\end{matrix}\right)(\mathbb ER_n^{}+o(1)) O\left(\left\{nh^p\right\}^{-\frac12}\right)+O(h^4)\\
        &=O\left(h^2\right)+O\left(n^{-1/2}h^{-1-p/2}\right)\to_p 0
    \end{split}
\end{equation}
where $R_n^{} = \frac1n \sum_{i=1}^n \left(\begin{matrix}1\\\frac{X_i-x}h\end{matrix}\right)K\left(\frac{X_i-x}h\right)[F(X_i)- m(X_i)]$ and $B(x,h)$ is defined in Theorem \ref{thm:oneorder-error}.
\end{theorem}

This theorem shows that the estimation using both $\mathcal L$ and $\mathcal U$ will not increase the order of expected error if $0< \beta <(6+p)^{-1}$.

By performing prediction-powered inference operations on local multivariable regression, the expected error of estimation with the given dataset remains $O(h^2)+O(n^{-1/2}h^{-1-p/2})$. Then, we shall give an analysis of the covariance of the estimations.

After obtaining the variance of $\widehat{\theta}^{\text{PP}}$, we have its asymptotic normality by

\begin{theorem}\label{thm:res_var}
    Under assumption of Theorem \ref{thm:estimation} and $F\in C^2(U)$, the local prediction-powered inference estimator $\widehat{\theta}^{\text{PP}}$ follows
\begin{equation}
     \left(\begin{matrix} \widehat{m(x)}^{\text{PP}}\\\widehat{\nabla m(x)}^{\text{PP}}  \end{matrix}\right)-\left(\begin{matrix} {m(x)} \\ \nabla m(x)   \end{matrix}\right) \to_d N\left(B(x,h)+O\left(h^4\right)+O\left(n^{-1/2}h^{2-p/2}\right), \text{Cov}\left(\widehat{\theta}^{\text{PP}}\right)\right).\\
\end{equation}
And there exist a constant $c_0$ that 
$$
\text{Cov}\left(\left(\begin{matrix} \widehat{m(x)}^{\text{con}}\\\widehat{\nabla m(x)}^{\text{con}}  \end{matrix}\right)\right)-\text{Cov}\left(\left(\begin{matrix} \widehat{m(x)}^{\text{PP}}\\\widehat{\nabla m(x)}^{\text{PP}}  \end{matrix}\right)\right)\succ n^{-1}h^{-2p}c_0\mathbf{I}
$$ where $c_0=\Omega(1)$ is defined in \ref{appsubsec:thm_res_var}. 
\end{theorem}

Here, we derive the asymptotic distribution of prediction-powered inference estimation with a Gaussian distribution. 
Compared with the covariance matrix for the estimation of conventional local multivariable regression, $\widehat{\theta}^{\text{con}}$, the covariance matrix of $\widehat{\theta}^{\text{PP}}$ is strictly "smaller". 
More specifically, we have constant $c_l, c_h$ such that $n^{-1}h^{-2p} c_l I \preceq \text{Cov}(\widehat{\theta}^{\text{con}})\preceq n^{-1}h^{-2p} c_h I$ according to the analysis of the proof of Theorem \ref{thm:res_var} in \ref{appsubsec:thm_res_var}.
And Theorem \ref{thm:res_var} shows that our prediction-powered inference approach decreases the covariance in a $O(1)$ constant ratio.

This positive definite difference shows that we reckon that the covariance of $\widehat\theta^{\text{PP}}$ is smaller than $\widehat\theta^{\text{con}}$ and local prediction-powered inference has a more stable estimation than conventional local multivariable regression.

Further analysis of the construction and comparison of the confidence interval/set is followed in the next subsection.

\subsection{Confidence Interval and Hypothesis Test}

Suppose that we have the estimation $\widehat\theta^{\text{PP}}$ with covariance matrix $\text{Cov}(\widehat\theta^{\text{PP}})=\{\sigma_{i,j}\}_{n\times n}$.
In traditional prediction-powered inference, the confidence interval and hypothesis test are set on separate components of the parameters. 

Define $\mathcal C_{i,\alpha}^{\text{PP}}$ as the confidence interval of $i$-th component of $\theta^*$ with expression
\begin{equation*}
    \mathcal C_{i,\alpha}^{\text{PP}} = \left[\widehat{\theta}^{\text{PP}} - z_{1-\alpha/2} \sigma_{i,i}, \widehat{\theta}^{\text{PP}} + z_{1-\alpha/2} \sigma_{i,i}\right]
\end{equation*}
where $z_{1-\delta}$ denotes the $\delta$ quantile of the standard normal distribution, for $\delta\in(0,1)$.
Such confidence $\mathcal C_{i,\alpha}^{\text{PP}}$ satisfies $\mathbb P(\theta^*_i\in \mathcal C_{i,\alpha}^{\text{PP}})\geq 1-\alpha$.
Since the function value $m(x)$ is the scalar value we want to estimate, and the estimation error and variance of function value and gradient value are of different magnitude (difference in the order of $h$), we can do t-test and construct one-dimensional confidence interval for $\widehat{m(x)}$:
\begin{equation}\label{equ:CIone}
    \mathcal C_{1,\alpha}^{\text{PP}} = \left[\widehat{m(x)}^{\text{PP}} - z_{1-\alpha/2} \cdot\text{S.E.}\left(\widehat{m(x)}^{\text{PP}}\right), \widehat{m(x)}^{\text{PP}} + z_{1-\alpha/2} \cdot \text{S.E.}\left(\widehat{m(x)}^{\text{PP}}\right)\right]
\end{equation}
and use chi-squared test to test and construct CI for the estimation of gradient $\widehat{\nabla m(x)}$.

By the conclusion of Theorem \ref{thm:res_var}, $\text{Cov}(\widehat{\theta}^{\text{con}})-\text{Cov}(\widehat{\theta}^{\text{PP}})\succ 0$ implies that each diagonal element $\sigma_{i,i}$ of $\text{Cov}(\widehat{\theta}^{\text{con}})$ is larger than that of $\text{Cov}(\widehat{\theta}^{\text{PP}})$. 
In other words, the confidence interval for estimation $\widehat{m(x)}^{\text{PP}}$ is smaller than that of $\widehat{m(x)}^{\text{con}}$, and consequently more has a more stable estimation.

Then we focus on the confidence region of the estimation of $\nabla m(x)$. By the asymptotic normality of estimation, the Hotelling's T-square distribution derives the confidence region as
\begin{equation}\label{equ:CImulti}
    \mathcal C_{2:p+1, \alpha}^{\text{DB}} = \left\{\nabla m(x) \bigg| \left( \widehat{\nabla m(x)}^{\text{PP}}-\nabla m(x)\right)^T\text{Cov}\left(\widehat{\nabla m(x)}^{\text{PP}}\right)^{-1}\left(\widehat{\nabla m(x)}^{\text{PP}}-\nabla m(x)\right)\leq \chi_{p}^2(1-\alpha)\right\}
\end{equation}
where $\chi_{k}^2(1-\alpha)$ is the $1-\alpha$ quantile of chi-square distribution with freedom of $k$.

The corresponding expressions for the confidence interval in $\widehat{m(x)}^{\text{con}}$ and the confidence region in $\widehat{\nabla m(x)}^{\text{con}}$ can be described in a form similar to Equation (\ref{equ:CIone}) and Equation (\ref{equ:CImulti}). 
We thus derive the conclusion for the deduced length of the confidence interval (volume of the confidence region) as follows:

\begin{theorem}\label{thm:CIdeduction}
    Under assumption of Theorem \ref{thm:estimation}, the length of the confidence interval $\mathcal C_{1,\alpha}^{\text{PP}}$, as indicated in Equation (\ref{equ:CIone}), is shorter than that of $\mathcal C_{1,\alpha}^{\text{con}}$ under similar circumstances; similarly, the volume of $\mathcal C_{2:p+1,\alpha}^{\text{PP}}$ in Equation (\ref{equ:CImulti}) is smaller than that of $\mathcal C_{2:p+1,\alpha}^{\text{con}}$ under corresponding conditions.
\end{theorem}

The deduction of the length of confidence interval and the volume of confidence region shows that the (co)variance of estimation $\widehat{m(x)}$ and $\widehat{\nabla m(x)}$ is decreased and consequently decreases the volatility via the implementation of prediction-powered inference on the local multivariable regression.
Together with Theorem \ref{thm:oneorder-error} and Theorem \ref{thm:estimation}, we show that the prediction-powered inference implement can decrease the variance without leading to a higher expected error.

Then, we need to prove the effectiveness of the confidence interval and region, that is, the probability that the ground truth value resides within the confidence set.

\subsection{Coverage Probability and Bias Correction}

Based on the analysis in advance, we have that the conventional local multivariable estimator $\widehat{\theta}^{\text{con}}$ and the local prediction-powered estimator $\widehat{\theta}^{\text{PP}}$ both follow the corresponding asymptotic normal distribution. 
The coverage probability of the confidence set of single variable and multivariable follows the following theorem.

\begin{theorem}\label{thm:coverageprob}
    Under assumption and region constructions of Theorem \ref{thm:res_var}, for single variable, that is, the coverage probability of biased confidence interval Equation \eqref{equ:CIone} with respect to function value $m(x)$ is 
    \begin{equation}\label{equ:CPone}
        \mathbb P\left\{m(x)\in \mathcal C_{1, \alpha}^{\text{PP}}\right\} = (1-\alpha)\left(1-\frac{h^4}{8\sigma_{1,1}^2}B_1^2(x)+O\left(h^6\right)+O\left(n^{-1/2}h^{2-p/2}\right)\right)
    \end{equation}
    where $B_1(x)=f(x)\mu_2\text{Tr}(\nabla^2m(x))$.
    When construct a bias correction confidence interval
    \begin{equation*}
        \mathcal C_{1,\alpha}^{\text{BC}} = \left[\widehat{m(x)}^{\text{PP}} -h^2 B_1(x)  - z_{1-\alpha/2} \cdot\text{S.E.}\left(\widehat{m(x)}^{\text{PP}}\right), \widehat{m(x)}^{\text{PP}} -h^2B_1(x) + z_{1-\alpha/2} \cdot \text{S.E.}\left(\widehat{m(x)}^{\text{PP}}\right)\right],
    \end{equation*}
    the coverage probability becomes $(1-\alpha)(1+O(h^6)+O(n^{-1/2}h^{2-p/2}))$, say, the error of coverage probability of confidence interval decreases from $O(h^4)$ to $O(h^6)$ if we apply bias correction when $0<\beta<(p+8)^{-1}$.
    
    For the multivariable condition, the coverage probability of Equation (\ref{equ:CImulti}) with respect to $\nabla m(x)$ is
    \begin{equation}\label{equ:CPmulti}
        \mathbb P\left\{{\nabla m(x)}\in \mathcal C_{2:p+1, \alpha}^{\text{PP}}\right\} = (1-\alpha)\left(1+\left(\frac12-c_1\right)\sum_{i=1}^p b_i^2+O(\widetilde h^3)\right).
    \end{equation}
    where $c_1 = \int_{\chi_p^2(1-\alpha)}^{\chi_{p+2}^2(1-\alpha)}\frac{e^{-y/2}y^{(2+p)/2-1}}{2^{(p+2)/2}\Gamma((p+2)/2)}dy$ is a given constant related to $p$, $\{b_i,i\in[p]\} =\text{Cov}(\widehat{\nabla m(x)})^{-1/2}B_2(x)$, $B_2(x)=(\frac{h^2}{2\mu_2f(x)}b_1(m)+\frac{h^2}{6\mu_2}b(m))$ and $\widetilde h = n^{-1/2}h^{1-p/2}$.
    When construct a bias correction confidence set
    \begin{equation*}
        \begin{split}
            \mathcal C_{2:p+1, \alpha}^{\text{BC}} = &\left\{\nabla m(x) \bigg| \left( \widehat{\nabla m(x)}^{\text{PP}}-\nabla m(x)-B_2(x)\right)^T\right.\\
            &\left.\cdot\text{Cov}\left(\widehat{\nabla m(x)}^{\text{PP}}\right)^{-1}\left(\widehat{\nabla m(x)}^{\text{PP}}-\nabla m(x)-B_2(x)\right)\leq \chi_{p}^2(1-\alpha)\right\}
        \end{split}
    \end{equation*}
    the coverage probability becomes $(1-\alpha)(1+O(\tilde h^3))$, say, the error of coverage probability of confidence set decreases from from $O(\tilde h^2)$ to $O(\tilde h^3)$ if we apply bias correction when $0<\beta<(p-2)^{-1}$.
    
\end{theorem}

Theorem \ref{thm:coverageprob} demonstrates that both the single-variable confidence interval and the multivariable confidence set of biased normality encompass the true values within the same order of the specified probability $(1-\alpha)$ with error orders of $O(h^4)$ and $O(n^{-1}h^{2-p})$, respectively. 
Furthermore, applying the bias correction operation would markedly reduce the error orders to $O(h^6)$ and $O(n^{-3/2}h^{3-3p/2})$, respectively.


Thus far, our analysis has shown that the coverage of confidence interval and set presented in $\mathcal C_1^{\text{PP}}$ and $\mathcal C_{2:p+2}^{\text{PP}}$ would converge to the theoretical target $1-\alpha$, with higher order of error if we implement bias correction approaches, showed in $\mathcal C_1^{\text{BC}}$ and $\mathcal C_{2:p+2}^{\text{BC}}$.

\subsection{High Dimensional Condition and Limited-sample Condition}

Another advantage of local prediction-powered inference in contrast of the conventional approach is to tackle with the relatively high dimensional condition and limited-sample condition.

In the conventional situation, when the dimension of the feature space is relatively high, the matrix $\mathbf{X}\mathbf{W}\mathbf{X}$ may be singular and consequently irreversible due to the weighting operation on the sparse sample space. 
More specifically, suppose that the domain of features is $p$-dimensional rectangular within $[-5, 5]$, for example.
Then the weight function $K(u)$ is positive if and only if $\|u\|_\infty \leq 1$.
After a calculation, we conclude that only $5^{-p}n$ samples can be calculated if the samples are distributed uniformly. 
The condition that $5^{-p}n < p$ can be considered as the lack of samples as well.
Even if the weight function can be set positive globally, the accuracy of computing program will cause this problem as well.

Recall the estimation of the labeled dataset $\widehat{\theta}^{\text{con}}=(\mathbf{X}\mathbf{W}\mathbf{X}^T)^{-1}\mathbf{X}\mathbf{W}\mathbf{Y}$. 
This estimation becomes intractable when the dimensionality is relatively high or the sample size is relatively limited, as $\mathbf{X}\mathbf{W}\mathbf{X}^T$ may exhibit singularity.
Estimator $\widehat{\theta}^{\text{PP}}=(\mathbf{\widetilde X}\mathbf{\widetilde W}\mathbf{\widetilde X}^T)^{-1}\mathbf{\widetilde X}\mathbf{\widetilde W}\mathbf{\widetilde Y}_F - (\mathbf{X}\mathbf{W}\mathbf{X}^T)^{-1}\mathbf{X}\mathbf{W}(\mathbf{Y}_F-\mathbf{Y})$ also fails for the same reason. Then we should find a substitution for $\widehat\Delta_{(n)}$, using both the response values of $\mathcal L$ and the features of $\mathcal U$.

Intuitively, we can replace $(\mathbf{X}\mathbf{W}\mathbf{X}^T)^{-1}$ by its expectation $n^{-1}(\mathbb E K((X_1-x)/h)X_1^+X_1^{+T})^{-1}$, while the latter expression can be estimated by $(1 + tN/n) ( \mathbf{X}\mathbf{W}\mathbf{X}^T+t\mathbf{\widetilde X}\mathbf{\widetilde W}\mathbf{\widetilde X}^T)^{-1}$, which is a non-singular matrix.
While $tN/n\to 0$, then this estimation converges to $n^{-1}(\mathbb E K_iX_i^+X_i^{+T})^{-1}$ while another estimation of rectifier follows
\begin{equation}
    \widehat\Delta^{\text{HD}}(t) = (1 + tN/n) ( \mathbf{X}\mathbf{W}\mathbf{X}^T+t\mathbf{\widetilde X}\mathbf{\widetilde W}\mathbf{\widetilde X}^T)^{-1}\mathbf{X}\mathbf{W}(\mathbf{Y}_F-\mathbf{Y})
\end{equation}
where the superscript $\text{HD}$ stands for high-dimensional.

\begin{theorem}\label{thm:high_dimensional}
    $\widehat{\Delta}^{\text{HD}}(t)$ is an unbiased estimator of $\mathbb E\Delta=(\mathbb EK_1X_1X_1^T)^{-1}\mathbb EK_1X_1(F(X_1)-Y_1)$. 
    Consequently, the estimator of high dimensional form $\widehat{\theta}^{\text{HD}}(t) =  (\widetilde{\mathbf{X}}\widetilde{\mathbf{W}}\widetilde{\mathbf{X}}^T)^{-1}\widetilde{\mathbf{X}}\widetilde{\mathbf{W}}\widetilde{\mathbf{Y}}_F - \widehat{\Delta}^{\text{HD}}(t)$, still has the same properties as the estimator $\widehat{\theta}^{\text{con}}$ and $\widehat{\theta}^{\text{PP}}$.
\end{theorem} 

When $t\to 0$, then the estimation of rectifier $\widehat{\Delta}^{\text{HD}}(t)$ converges to $\widehat{\Delta}$ in $\widehat{\theta}^{\text{PP}}$, which maintain the invertibility, low-volatility and other superiority of the estimator of local prediction-powered inference.

\section{Experiments}\label{sec:exp}
In this section, numerical simulations and real-data experiments are conducted to demonstrate the priority proposed in Section \ref{sec:theo}. Numerical simulations, which generate data within a specified piecewise function, help to establish the universality of local multivariable regression in estimating a particular target point. Real-data on house prices, which is apt for using local PPI owing to its characteristics, further illustrates that our method can achieve more stable volatility. All the Python code is available at \url{https://github.com/yanwugu2001/Local-Prediction-Powered-Inference}.

\subsection{Numerical Simulation}

In the numerical simulation, we introduce a piecewise function with covariate $X\in\mathbb R^{10}$. 
Different components give different contributions to the function value $m(x)$. 
Specifically, $m(x)=m_1(x_1, x_2) + m_2(x_3) + m_3(x_4, x_5,x_6, x_7)$  where
\begin{equation*}
    \begin{split}
        m_1(x_1, x_2) &=|x_1x_2|\\
        m_2(x_3) &=\begin{cases}
            x_3 \times \cos(\pi x_3), &x_3 \leq 0\\
            \sin(\pi x_3), &x_3 > 0
        \end{cases}\\
        m_3(x_4, x_5,x_6, x_7) &= - x_4 - 0.5 \times x_5 + 0.5 \times x_6 + x_7\\
    \end{split}
\end{equation*}
In this context, $x_8, x_9$, and $x_{10}$ are extraneous as they do not affect $Y$. 
The function $m(x)$ encompasses linear, non-linear, and stochastic influences of its components, with the piecewise nature causing shifts in both the function value and gradient, thereby introducing complexities in the estimation process.

To construct the feature $X$, we extract 100,000 instances, 10,000 instances, and 1,000,000 instances from the identical Gaussian distribution $N(\mathbf{0}, \mathbf{I}_{10})$ for the model training set, the labeled dataset $\mathcal{L}$, and the unlabeled dataset $\mathcal{U}$, respectively. 
For the associated function value $m(x)$, we introduce a noise component with a variance $\varepsilon_i$ of 0.2 to the label $Y$. 
Consequently, the overall variance of $Y$ is approximately 1.9.

For the good predictor $F$, we utilized the XGBoost algorithm, which was trained on a dataset consisting of 100,000 samples that are identically and independently distributed in relation to the datasets $\mathcal{L}$ and $\mathcal{U}$, to make sure its efficiency and independence on inference datasets.
This tree-based model achieves an approximate mean squared error (MSE) of $0.1$ on $\mathcal{L}$ and $\mathcal{U}$, thereby demonstrating its superiority.

To rigorously evaluate the efficacy of our local prediction-powered inference in comparison to conventional local multivariable regression, we employ the bootstrap methodology to estimate the error of estimation. 
Specifically, a single sample from the labeled dataset $\mathcal L$ is designated as the target point. 
The remaining $n=9,999$ samples are then utilized to perform local multivariable regression, yielding estimates of both the function value and its gradient. 
Subsequently, local prediction-powered inference is conducted on the identical target point, this time utilizing the unlabeled dataset $\mathcal U$. 
This procedure is iteratively applied to 1,000 random samples from the 10,000 labeled instances, resulting in the computation of the mean squared error for both function value and gradient estimations.

Regarding the selection of the kernel function $K(\cdot)$ and the hyperparameter $h$, we have empirically determined $K(x) = (2\pi)^{-p/2}\exp\{-\|x\|_2^2/2\}$ and set $h$ at 0.5. 
The configuration parameters of the tree model include a total of 300 trees, a maximum depth of 8 per tree, a maximum of 128 leaves per tree, and a learning rate of 0.1.

The error scatter result of numerical experiments are plotted as Figure \ref{fig:error_scatter}:
\begin{figure}[ht]
    \centering
    \includegraphics[width=\linewidth]{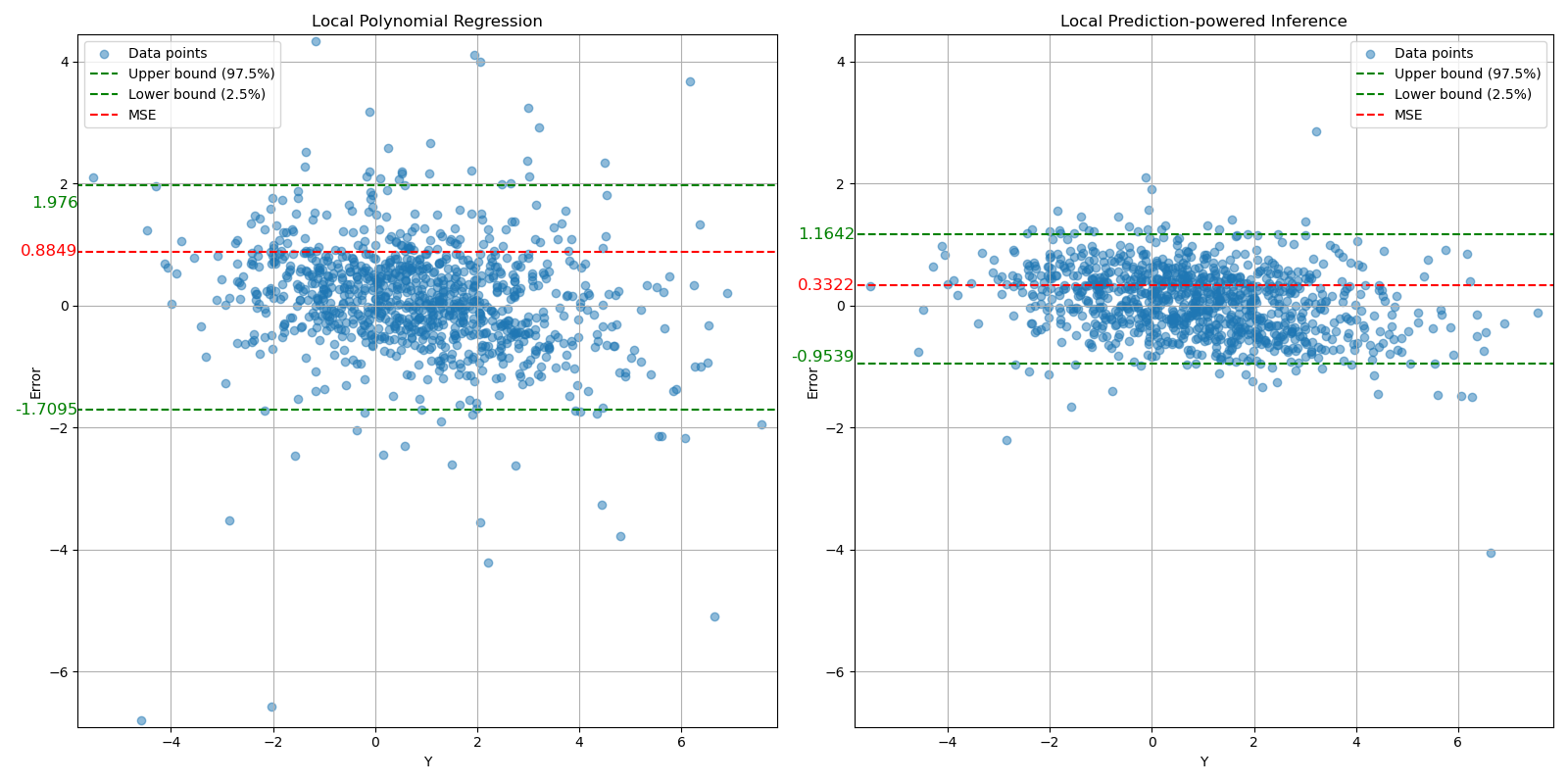}
    \caption{Error Scatter Plot}
    \label{fig:error_scatter}
\end{figure}

In this Y-error scatter plot, the upper bound at the 97.5\% quantile and the lower bound at the 2.5\% quantile are depicted using green dashed lines. 
These quantile lines illustrate that local prediction-powered inference can effectively reduce the width of the confidence intervals from $[-1.71, 1.98]$ to $[-0.95, 1.16]$, thereby enhancing the precision of the inference by 43\%. 
Currently, the mean squared error (MSE) reduced by 62\%, indicated by a red dashed line, further substantiates this assertion.

\begin{figure}[ht]
    \centering
    \includegraphics[width=0.8\linewidth]{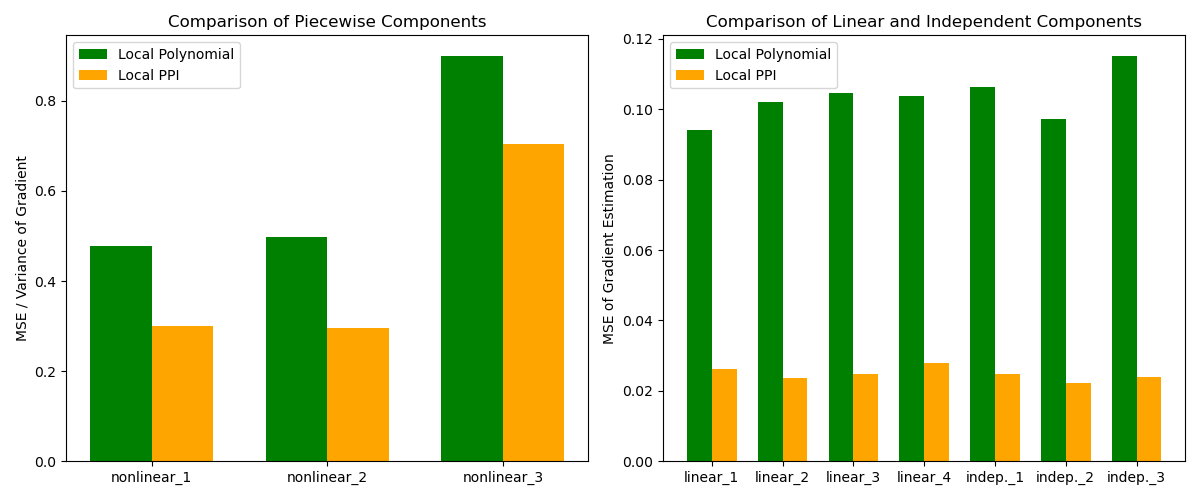}
    \caption{Error of Gradient Estimation}
    \label{fig:error_gradient}
\end{figure}

The gradient estimations illustrated in Figure \ref{fig:error_gradient} indicate that the local prediction-powered inference method can effectively decrease the MSE in gradient estimation. 

On the left side of Figure \ref{fig:error_gradient}, the standardized MSE, i.e., MSE divided by the standard error, of three non-linear piecewise components are depicted. 
Although estimating such gradients is challenging due to the potential distribution of sample instances in divergent directions of the target, leading to significant error and volatility, the local prediction-powered inference method consistently reduces the MSE, yielding a reduction range of 21\% to 40\%.
Conversely, the right subplot, which is based on linear and independent components, exhibits components with a globally invariant gradient value. 
Within this inference framework, although traditional methodologies produce relatively adequate estimations, our proposed approach demonstrates a substantial enhancement, ranging from 70\% to 80\% improvement.

\begin{figure}[htbp]
    \centering
    \begin{subfigure}[b]{\textwidth}
        \centering
        \includegraphics[width=0.8\textwidth]{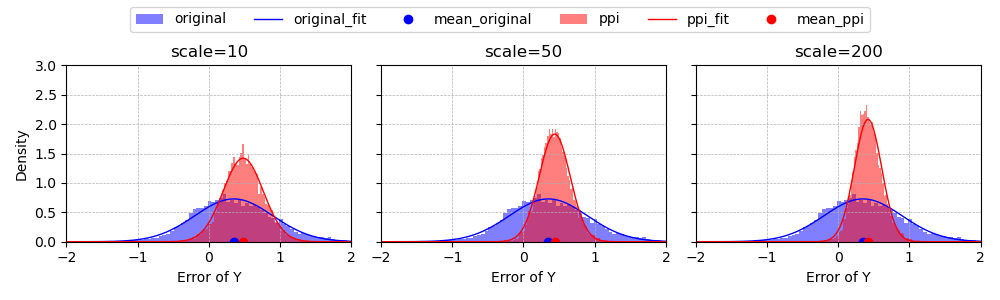}
        \caption{The Distribution of Function Value Error}
        \label{fig:subfig1}
    \end{subfigure}
    
    \begin{subfigure}[b]{\textwidth}
        \centering
        \includegraphics[width=0.8\textwidth]{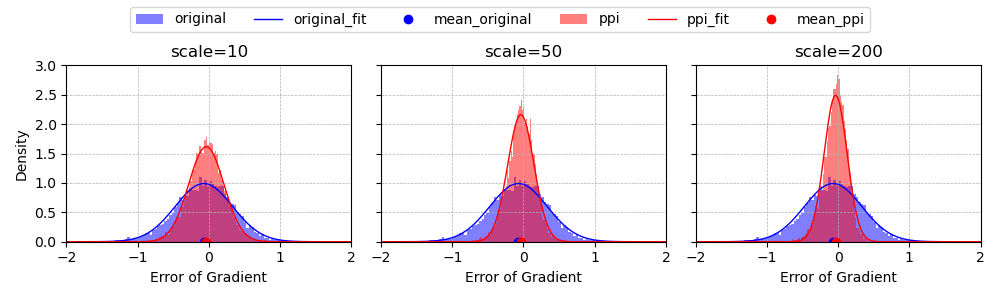}
        \caption{The Distribution of Gradient Value Error} 
        \label{fig:subfig3}
    \end{subfigure}
    
    \caption{Fitted Normal Distribution Comparison}
    \label{fig:distribution_comp}
\end{figure}

Figure \ref{fig:distribution_comp} presents the bar charts depicting the Density-Error of the response variable $Y$ and gradient. 
In both subplots, various scales of the size of unlabeled and labeled datasets are tested, specifically at 10, 50, and 200. 
In accordance with Theorem \ref{thm:res_var}, it is possible to approximate a normal distribution to the density of the error and consequently plot a fitted probability density function. 
The figure demonstrates that local prediction-powered inference is capable of reducing the variance of the error while not significantly augmenting the error itself, thereby rendering the error distribution approximately normal.
Furthermore, an increase in the scale results in a further reduction in variance, which is consistent with our theorem.

For the variance and coverage probability of estimation via multivariable and local prediction-powered inference, we extract a decile of data instances from the labeled dataset $\mathcal{L}$ and the unlabeled dataset $\mathcal{U}$, conducting the inference operation at a fixed target point 100 times. 
Sequentially, we assess the variance of the estimated values and replicate the aforementioned operations for 1,000 different target points selected from the labeled dataset. 
Consequently, the coverage probability is derived from the ensuing simulation.
For simplicity, only one-dimensional confidence intervals are considered for the estimation of function values and gradients.

As illustrated in Table \ref{tab:CP_simulation}, in the absence of bias correction, the coverage probabilities span from 86.9\% to 92.8\% for local multivariable regression and from 88.4\% to 93.1\% for local prediction-powered inference. 
Upon applying the de-biasing technique to account for second-order errors, there is a significant improvement in the coverage probabilities, aligning them more closely with 95\%. 
The standard error reduction attributable to the prediction-powered inference mechanism exceeds 50\% in all combinations of dataset sizes. 
Furthermore, as the size of the dataset increases, the standard error is observed to decline, as demonstrated in our findings, without compromising the coverage probability as reported in $1-\alpha$.

\begin{table}[!ht]
    \centering
    \caption{The Coverage Probability of (De-)Biased Confidence Intervals}
    \begin{tabular}{|c|c|c|c|c|c|}
    \hline
    \makecell[c]{Dataset Size\\ $n$(, $N$)} & Method & \makecell[c]{Coverage\\ Probability(\%)}  & \makecell[c]{De-Biased Coverage\\ Probability(\%)} & Standard Error & S.E. Decay(\%)\\ \hline
    \multirow{2}*{100, 10000}& Local Multi.& 92.8 & 93.5 & 2.37 & \multirow{2}*{59.1} \\ 
    ~ & Local PPI & 92.1  & 94.2  & 0.97& ~ \\ \hline
    \multirow{2}*{200, 20000}& Local Multi.& 92.9 & 94.1 & 1.88 & \multirow{2}*{53.2}  \\ 
    ~ & Local PPI & 93.1 & 94.3  & 0.80& ~  \\ \hline
    \multirow{2}*{500, 50000} & Local Multi. & 88.1 & 92.3  & 1.16 &   \multirow{2}*{66.3}\\ 
    ~& Local PPI & 88.9 &94.4  & 0.56 &~ \\ \hline
    \multirow{2}*{1000, 100000}& Local Multi. & 88.2 & 94.2  & 1.05 & \multirow{2}*{52.4}  \\ 
    ~& Local PPI & 88.4 & 94.1  & 0.50 & ~\\ \hline
    \multirow{2}*{2000, 200000} & Local Multi. & 86.8 & 91.8 & 0.82  &\multirow{2}*{54.9} \\ 
    ~& Local PPI  & 91.1 & 93.2  & 0.45 &~\\ \hline
    \end{tabular}
    
    \label{tab:CP_simulation}
\end{table}

\subsection{House Price Inference}

The prediction of house price is always an essential regression problem.
The dataset employed for forecasting the sales prices of residential properties in King County is sourced from Kaggle. 
This dataset covers 21,613 instances, each annotated with 20 distinct attributes of houses alongside the corresponding sale prices, covering transactions executed from May 2014 to May 2015.
The features of such regression problem include:
\begin{itemize}
    \item The size and room numbers of the house.
    \item The year the house was built and renovated.
    \item The quality of the house and the facilities.
    \item The location and view of the house.
\end{itemize}
The first two types of terms are objective, while the others are subjective and graded by some property assessors. 

Among the 20 attributes, six are continuous numerical variables that quantify the spatial dimensions and geographical coordinates of the property. 
These continuous variables provide an essential overview of the structural characteristics of the home and relevant information. 
We decompose these variables into two primary components via Principal Component Analysis (PCA). 
The remainder of the attributes are discrete variables that offer more detailed information on aspects such as construction year, number of rooms, presence on the waterfront, and subjective scores. We aggregate the construction and renovation years into a single principal component and further decompose the remaining discrete (yet ordinal) and objective variables into two principal components.
Alongside the evaluation scores, "grade" and "condition", we employ these seven features for local multivariable regression and prediction-powered inference techniques to assess the impact of implementation.

By dropping several NaN (Not a Number) data, 21597 instances are left, and we split them into train dataset (10,000 instances), labeled dataset (1,500 instances), unlabeled dataset (10,000 instances) and test dataset (97 instances).
Under the paradigm of prediction-powered inference, we trained a model under a train dataset and gave predictions to the labeled dataset and the unlabeled dataset.
For the sample size of (un)labeled dataset, we tested with same ratio 10 for four times: (100, 1000), (200, 2000), (400, 4000), (800, 8000) and with fixed labeled size 100 for four times: (100, 1000), (100, 2000), (100, 4000), (100, 8000). The mean absolute error (MAE) and the standard error of the estimations are contrasted.

\begin{figure}[htbp]
    \centering
    \includegraphics[width=\textwidth]{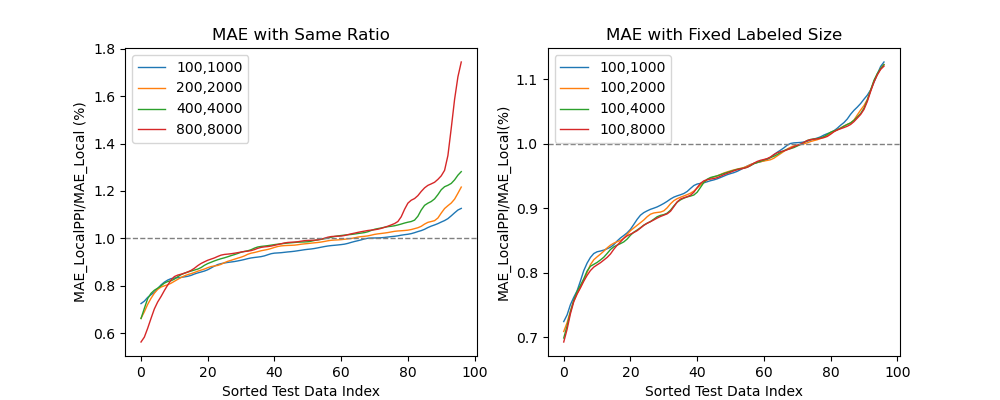}
    \caption{Deduction of Mean Absolute Error}
    \label{fig:MAE_price}
\end{figure}

\begin{figure}[htbp]
    \centering
    \includegraphics[width=\textwidth]{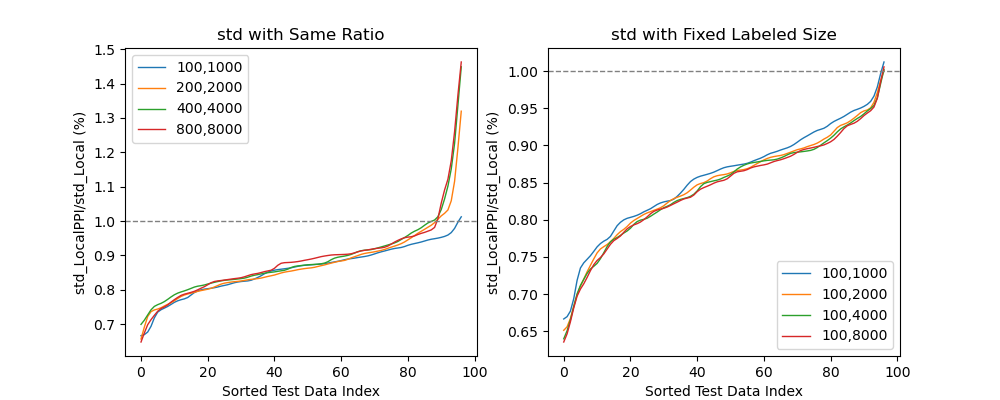}
    \caption{Deduction of Standard Error of Estimation}
    \label{fig:std_price}
\end{figure}

For the 97 instances within the test dataset, each instance was designated as the target point. 
The corresponding segments of the labeled and unlabeled datasets were sampled, followed by performing inference 100 times under sampling data. 
The mean absolute error (MAE) and standard error (S.E.) for each target point were calculated. 
Subsequently, the 97 instances in the test dataset were ranked and the deduction of MAE and SE under the prediction model were compared, as illustrated in Figures \ref{fig:MAE_price} and \ref{fig:std_price}.

From Figure \ref{fig:MAE_price}, it is observed that for each target data point, the mean absolute errors (MAEs) of local multivariable inference and local PPI exhibit comparable performance. 
An increment in the size of the unlabeled dataset results in a marginal increase in the absolute error. This escalation in the performance of MAEs, despite maintaining a constant unlabeled-labeled ratio, can probably be attributed to the suboptimal performance of the underlying model.

In Figure \ref{fig:std_price}, more than 90\% of the instances demonstrate an improvement in variance performance when maintaining a fixed unlabeled-labeled ratio. 
Additionally, nearly all instances exhibit more stable estimations with a fixed labeled dataset size under the technique of prediction-powered inference. 
Furthermore, an increase in the unlabeled dataset size, while keeping the labeled dataset size constant, augments the stability of performance as corroborated by theoretical analysis.
Actually, the Mean Squared Error (MSE) of the same target point has the same expression as the standard error, which has the same conclusion.

It is worth mentioning that due to the lack of training set, the XGBoost predictor $F$ still suffers from a relatively high error. But this shortcoming does not significantly influence the performance of prediction-powered inference, because of the debias operations taken by rectifier $\Delta$.

In conclusion, compared to the local multivariable regression, the local prediction-powered inference can give a more stable estimation without the higher cost of absolute error.

\subsection{Air Quality Inference}\label{subsec:air_quality}

In this real-data experiment, we focus on a dataset of hourly air quality in India. 
Twenty monitoring stations give over 200,000 records from 2015 to 2020 with 19 observable variables including particulate pollutants, nitrogen oxides, combustion gaseous pollutants, sulfur compounds and volatile organic compounds.
The AQI (Air Quality Index) has a complex calculation method related to the above variables which can be recognized as the potential function $m(x)$. Thus, our target is to use various pollutant contents to estimate AQI.

The data were compiled from the website of the Central Pollution Control Board (CPCB) \url{https://cpcb.nic.in/}, the official authority of the Government of India, and the complete version is available on the Kaggle website \url{https://www.kaggle.com/datasets/rohanrao/air-quality-data-in-india}.

In our configuration, the dataset is partitioned into training, testing, labeled, and unlabeled subsets. 
The test dataset comprises records from a particular station, kept aside for evaluation purposes. 
The training dataset encompasses numerous records that may contain missing data, which, although not useful for inference, contribute to the effective training of the XGBoost model. 
The unlabeled dataset includes records where some equipment may have incomplete or inaccurate data, resulting in the absence of labels. 
Conversely, the labeled dataset contains complete and accurate data required for the implementation of the local PPI method.

\begin{figure}[htb]
\centering
\subfloat[Detailed PPI Improvement]{\includegraphics[width=0.48\linewidth]{pictures/air_quality_arrow.png}\label{fig:air_quality_arrow}}
\hfill
\subfloat[Overall Std and MSE Comparison]{\includegraphics[width=0.48\linewidth]{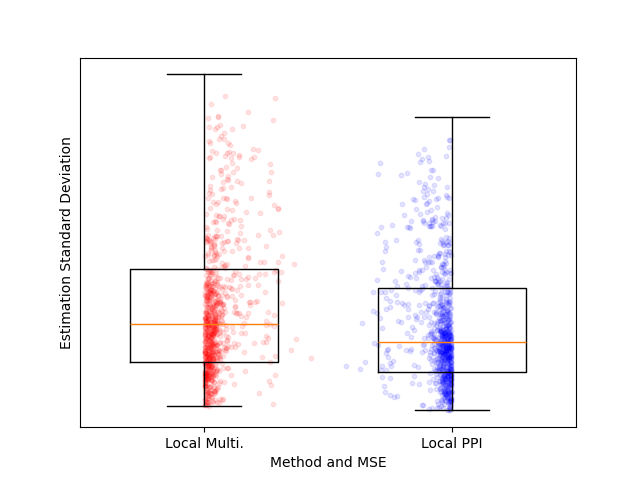}\label{fig:air_quality_box}}
\caption{Comparison of Local Multivariable Regression and Local PPI}
\label{fig:air_quality_comparison}
\end{figure}

For each target point in the test dataset, we perform bootstrapping 100 times by sampling 2,000 instances from the labeled dataset and 20,000 instances from the unlabeled dataset, respectively. 
Applying the bandwidth $h=0.7$ and PCA operations to the pollution content clusters, local prediction-powered inference based on the 5-variable  demonstrates superior performance as Figure \ref{fig:air_quality_comparison}. 

Figure \ref{fig:air_quality_arrow} presents a comparison between local PPI and local multivariable inference for each target point. 
When the arrows point to the right, the PPI method reduces the standard deviation of the estimation for the respective target point; and the arrows pointing upward indicate that the PPI method decreases the mean squared error of the estimation for the corresponding target point. 
Conversely, directions towards the left and downward signify high volatility and low accuracy.
The overall statistic of arrow plot is listed before in Figure \ref{fig:air_sub_pie}.

Figure \ref{fig:air_quality_box} presents a comparison between local multivariable inference and local prediction-powered inference, using a combination of box plots and scatter plots to illustrate estimation standard deviation and mean squared error (MSE). 
The box plots reveal that the method on the right has a lower median standard deviation and a more compact interquartile range, indicating a reduction in variability compared to the method on the left. 
This suggests a more consistent performance in the estimations. 
The scatter plots, where the height of each point corresponds to its standard deviation and the horizontal distance from the box plot’s central line reflects its MSE, show that despite the reduced variance in the second method, there is no noticeable increase in error. 
Both methods maintain similar distributions of MSE, with points scattered relatively evenly around the central line. In general, the method on the right demonstrates improved stability by reducing the variance of the estimate without introducing higher errors, making it more effective in maintaining accuracy.

Broadly speaking, our innovative approach harnesses the power of the unlabeled dataset in conjunction with models adeptly trained on missing-feature data. 
This synergy not only significantly bolsters the predictive prowess of local multivariable inference, enhancing its stability to a remarkable degree, but does so without ever sacrificing accuracy.

\section{Conclusions}\label{sec:con}
The simulation experiment and the real data trial proved that local prediction-powered inference can reduce volatility of the estimation, especially when the sample size of the labeled dataset is limited. 

In contrast to the evaluation of traditional inference methodologies, our analysis focuses on the theoretical performance at a specific target point, i.e., locally rather than across global conditions.
Given an expected error of equal equality, the predictor $F$ demonstrably yields a lower variance, as substantiated by theoretical proof. 
Furthermore, the confidence interval retains the same order of magnitude irrespective of bias adjustment.
Coverage probabilities are validated via elementary algebra in one dimension and through the application of an introduced biased Beta distribution in multiple dimensions.

The improvement of local prediction-powered inference in contrast of simply applying prediction-powered inference, includes:
\begin{itemize}
    \item The computation of (sub)gradients of PPI is replaced by explicit solution expressed by the matrix of features, weights and response values, which improves the computation efficiency;
    \item The dependence of components can be described by the inverse of matrix of features, in contrast of the independence of classical PPI approach.
\end{itemize}

There are also several open problems of prediction-powered inference technique, including:
\begin{itemize}
    \item The criterion of good predictor $F$ which to determine whether use the PPI or not;
    \item The general paradigm of PPI;
    \item The implementation of other non-linear and no-explicit-solution optimization problem.
\end{itemize}

Notwithstanding, local prediction-powered inference offers a methodology to enhance the stability of estimations for a specified local target. 
Despite the constraints in the size of the labeled dataset, our approach remains effective. 
Furthermore, local prediction-powered inference can be employed in high-cost design scenarios with commendable simulation techniques, or in social investigation issues that can be addressed through alternative investments.

\section*{Acknowledgments}
Yanwu Gu's research was partially supported by HKPFS PF22-69747. Dong Xia's research was partially supported by Hong Kong RGC grant GRF 16300121.

\bibliographystyle{unsrt}  
\bibliography{references}  

\begin{thebibliography}{10}

\bibitem{savitzky1964smoothing}
Abraham Savitzky and Marcel~JE Golay.
\newblock Smoothing and differentiation of data by simplified least squares procedures.
\newblock {\em Analytical chemistry}, 36(8):1627--1639, 1964.

\bibitem{cleveland1979robust}
William~S Cleveland.
\newblock Robust locally weighted regression and smoothing scatterplots.
\newblock {\em Journal of the American statistical association}, 74(368):829--836, 1979.

\bibitem{cleveland1988locally}
William~S Cleveland and Susan~J Devlin.
\newblock Locally weighted regression: an approach to regression analysis by local fitting.
\newblock {\em Journal of the American statistical association}, 83(403):596--610, 1988.

\bibitem{lin2017feature}
Cheng-Kuan Lin and Heiu-Jou Shaw.
\newblock Feature-based estimation of preliminary costs in shipbuilding.
\newblock {\em Ocean Engineering}, 144:305--319, 2017.

\bibitem{lu1996multivariate}
Zhan-Qian Lu.
\newblock Multivariate locally weighted polynomial fitting and partial derivative estimation.
\newblock {\em journal of multivariate analysis}, 59(2):187--205, 1996.

\bibitem{angelopoulos2023prediction}
Anastasios~N. Angelopoulos, Stephen Bates, Clara Fannjiang, Michael~I. Jordan, and Tijana Zrnic.
\newblock Prediction-powered inference.
\newblock {\em Science}, 382(6671):669--674, 2023.

\bibitem{loader2006local}
Clive Loader.
\newblock {\em Local regression and likelihood}.
\newblock Springer Science \& Business Media, 2006.

\bibitem{fan1996study}
Jianqing Fan, Ir{\`e}ne Gijbels, Tien-Chung Hu, and Li-Shan Huang.
\newblock A study of variable bandwidth selection for local polynomial regression.
\newblock {\em Statistica Sinica}, pages 113--127, 1996.

\bibitem{gunnarsson2024prediction}
Elias~S{\o}vik Gunnarsson, H{\aa}kon~Ramon Isern, Aristidis Kaloudis, Morten Risstad, Benjamin Vigdel, and Sjur Westgaard.
\newblock Prediction of realized volatility and implied volatility indices using ai and machine learning: A review.
\newblock {\em International Review of Financial Analysis}, page 103221, 2024.

\bibitem{cai2021unified}
Mingxuan Cai, Jiashun Xiao, Shunkang Zhang, Xiang Wan, Hongyu Zhao, Gang Chen, and Can Yang.
\newblock A unified framework for cross-population trait prediction by leveraging the genetic correlation of polygenic traits.
\newblock {\em The American Journal of Human Genetics}, 108(4):632--655, 2021.

\bibitem{nadaraya1964estimating}
Elizbar~A Nadaraya.
\newblock On estimating regression.
\newblock {\em Theory of Probability \& Its Applications}, 9(1):141--142, 1964.

\bibitem{watson1964smooth}
Geoffrey~S Watson.
\newblock Smooth regression analysis.
\newblock {\em Sankhy{\=a}: The Indian Journal of Statistics, Series A}, pages 359--372, 1964.

\bibitem{gasser1979kernel}
Theo Gasser and Hans-Georg M{\"u}ller.
\newblock Kernel estimation of regression functions.
\newblock In {\em Smoothing Techniques for Curve Estimation: Proceedings of a Workshop held in Heidelberg, April 2--4, 1979}, pages 23--68. Springer, 1979.

\bibitem{fan1993local}
Jianqing Fan.
\newblock Local linear regression smoothers and their minimax efficiencies.
\newblock {\em The annals of Statistics}, pages 196--216, 1993.

\bibitem{gasser1985kernels}
Theo Gasser, Hans-Georg Muller, and Volker Mammitzsch.
\newblock Kernels for nonparametric curve estimation.
\newblock {\em Journal of the Royal Statistical Society. Series B (Methodological)}, pages 238--252, 1985.

\bibitem{fan1997local}
Jianqing Fan, Theo Gasser, Ir{\`e}ne Gijbels, Michael Brockmann, and Joachim Engel.
\newblock Local polynomial regression: Optimal kernels and asymptotic minimax efficiency.
\newblock {\em Annals of the Institute of Statistical Mathematics}, 49:79--99, 1997.

\bibitem{fan1995data}
Jianqing Fan and Irene Gijbels.
\newblock Data-driven bandwidth selection in local polynomial fitting: variable bandwidth and spatial adaptation.
\newblock {\em Journal of the Royal Statistical Society: Series B (Methodological)}, 57(2):371--394, 1995.

\bibitem{ruppert1995effective}
David Ruppert, Simon~J Sheather, and Matthew~P Wand.
\newblock An effective bandwidth selector for local least squares regression.
\newblock {\em Journal of the American Statistical Association}, 90(432):1257--1270, 1995.

\bibitem{wang2020methods}
Siruo Wang, Tyler~H McCormick, and Jeffrey~T Leek.
\newblock Methods for correcting inference based on outcomes predicted by machine learning.
\newblock {\em Proceedings of the National Academy of Sciences}, 117(48):30266--30275, 2020.

\bibitem{robins1994estimation}
James~M Robins, Andrea Rotnitzky, and Lue~Ping Zhao.
\newblock Estimation of regression coefficients when some regressors are not always observed.
\newblock {\em Journal of the American statistical Association}, 89(427):846--866, 1994.

\bibitem{Zhang2016SemisupervisedIG}
Anru~R. Zhang, Lawrence~D. Brown, and T.~Tony Cai.
\newblock Semi-supervised inference: General theory and estimation of means.
\newblock {\em The Annals of Statistics}, 2016.

\bibitem{chakrabortty2022semi}
Abhishek Chakrabortty, Guorong Dai, and Raymond~J Carroll.
\newblock Semi-supervised quantile estimation: Robust and efficient inference in high dimensional settings.
\newblock {\em arXiv preprint arXiv:2201.10208}, 2022.

\bibitem{azriel2022semi}
David Azriel, Lawrence~D Brown, Michael Sklar, Richard Berk, Andreas Buja, and Linda Zhao.
\newblock Semi-supervised linear regression.
\newblock {\em Journal of the American Statistical Association}, 117(540):2238--2251, 2022.

\bibitem{chakrabortty2018efficient}
Abhishek Chakrabortty and Tianxi Cai.
\newblock {Efficient and adaptive linear regression in semi-supervised settings}.
\newblock {\em The Annals of Statistics}, 46(4):1541 -- 1572, 2018.

\bibitem{zhang2022high}
Yuqian Zhang and Jelena Bradic.
\newblock High-dimensional semi-supervised learning: in search of optimal inference of the mean.
\newblock {\em Biometrika}, 109(2):387--403, 2022.

\bibitem{song2024general}
Shanshan Song, Yuanyuan Lin, and Yong Zhou.
\newblock A general m-estimation theory in semi-supervised framework.
\newblock {\em Journal of the American Statistical Association}, 119(546):1065--1075, 2024.

\bibitem{angelopoulos2023ppi++}
Anastasios~N Angelopoulos, John~C Duchi, and Tijana Zrnic.
\newblock Ppi++: Efficient prediction-powered inference.
\newblock {\em arXiv preprint arXiv:2311.01453}, 2023.

\bibitem{leek2007capturing}
Jeffrey~T Leek and John~D Storey.
\newblock Capturing heterogeneity in gene expression studies by surrogate variable analysis.
\newblock {\em PLoS genetics}, 3(9):e161, 2007.

\bibitem{risso2014normalization}
Davide Risso, John Ngai, Terence~P Speed, and Sandrine Dudoit.
\newblock Normalization of rna-seq data using factor analysis of control genes or samples.
\newblock {\em Nature biotechnology}, 32(9):896--902, 2014.

\bibitem{smyth2005limma}
Gordon~K Smyth.
\newblock Limma: linear models for microarray data.
\newblock In {\em Bioinformatics and computational biology solutions using R and Bioconductor}, pages 397--420. Springer, 2005.

\end{thebibliography}
\appendix
\section{Proof of Theorems}\label{appsec:proof}

\subsection{Proof of Theorem \ref{thm:oneorder-error}}\label{appsubsec:thm_oneorder_error}
\begin{proof}
    First, we decompose the expression of error as
    \begin{equation*}
        \begin{split}
            \mathbb E( \widehat{\theta}_{(n)}-\theta^*|X_1,\dots,X_n) &= \mathbb E( \widehat{\theta}_{(n)}|X_1,\dots,X_n)-\theta^*  \\
            &= (\mathbf X\mathbf W\mathbf X^T)^{-1}\mathbf X\mathbf W(\mathbf M-\mathbf X^T\beta)\\
            &= \text{diag}\{1,h^{-1}I_p\} S_n^{-1}R_n,
        \end{split}
    \end{equation*}
    where
    \begin{equation*}
        \begin{split}
            \mathbf M &= \left(\begin{matrix}m(X_1)&\cdots&m(X_n)\end{matrix}\right)^T,\\
            S_n &= \frac 1n\sum_{i=1}^nh^{-p}\left(\begin{matrix}1\\\frac{X_i-x}{h}\end{matrix}\right)\left(\begin{matrix}1&\frac{X_i-x}h\end{matrix}\right)K\left(\frac{X_i-x}h\right),\\
            R_n &= \frac 1n\sum_{i=1}^n\left(\begin{matrix}1\\ \frac{X_i-x}h\end{matrix}\right)\left[m(X_i)-m(x)-\nabla m^T(x)(X_i-x)\right]h^{-p}K\left(\frac{X_i-x}{h}\right).\\
        \end{split}
    \end{equation*}

    To estimate $S_n^{-1}$ and $R_n$, we use the Central Limit Theorem and we have 
    \begin{equation*}
        \begin{split}
            \mathbb ES_n&= \int h^{-p}\left(\begin{matrix}1\\\frac{X_1-x}{h}\end{matrix}\right)\left(\begin{matrix}1&\frac{X_1-x}h\end{matrix}\right)K\left(\frac{X_1-x}h\right)f(X_1)dX_1\\
            &=\int \left(\begin{matrix}1\\u\end{matrix}\right)\left(\begin{matrix}1&u\end{matrix}\right)K(u)f(x+hu)du:=A(h),\\
            \sqrt{nh^p}(S_n - A(h)) &= O_p(1),\\
            S_n &= A(h) + O_p(\{nh^p\}^{-1/2}).
        \end{split}
    \end{equation*}

    Since 
    \begin{equation*}
        S_n^{-1} = A^{-1}(h)+O_p(\{nh^p\}^{-1/2}),
    \end{equation*}
    and 
    \begin{equation*}
        \begin{split}
            A(h) &= \int\left( \begin{matrix} 1&u^T\\ u&uu^T \end{matrix} \right) K(u)f(x+hu)du\\
            &= f(x) \int\left( \begin{matrix}1&u^T\\u&uu^T\end{matrix} \right) K(u)du+h\int\left( \begin{matrix}1&u^T\\u&uu^T\end{matrix} \right) \nabla f(x)^TuK(u)du + O(h^2)\\
            &=\left(\begin{matrix}f(x) &h\mu_2\nabla f(x)^T\\h\mu_2\nabla f(x) & \mu_2f(x)I_p \end{matrix}\right)+ O(h^2),
        \end{split}
    \end{equation*}
    using the inverse matrix formula
    \begin{equation*}
        \left(\begin{matrix}A & B^T\\ B&D\end{matrix}\right)^{-1} = \left(\begin{matrix}E^{-1}&-E^{-1}F^T \\ -FE^{-1} & D^{-1}+FE^{-1}F^T \end{matrix}\right),
    \end{equation*}
    where $E = A-B^TD^{-1}B, F=D^{-1}B$, we have
    \begin{equation*}
        \begin{split}
            E&=f(x)-\frac{h^2\mu_2}{f(x)}\nabla f(x)^T\nabla f(x),\\
            F&= \frac{h}{f(x)} \nabla f(x).\\
        \end{split}
    \end{equation*}
    And then we conclude that
    \begin{equation}\label{equ:A_inv_h}
        \begin{split}
            A^{-1}(h) &= \left(\begin{matrix} E^{-1}&-E^{-1}F^T\\-FE^{-1} & D^{-1}+FE^{-1}F^T\end{matrix}\right)\\
            &=\left(\begin{matrix}1/f(x)+O(h^2) & -h /f^2(x)\cdot \nabla f(x)^T+O(h^3)\\-h /f^2(x)\cdot \nabla f(x) +O(h^3)& 1/(\mu_2 f(x))I+O(h^2)\\\end{matrix}\right)\\
            &=\frac1{f(x)}\left(\begin{matrix}1 & -h /f(x)\cdot \nabla f(x)^T\\-h /f(x)\cdot \nabla f(x) & 1/\mu_2 \cdot I_p\\\end{matrix}\right)+O(h^2).
        \end{split}
    \end{equation}

    For the residual term, using the Assumption \ref{asump:multi} (ii) to get 
    \begin{equation*}
        \begin{split}
            \mathbb ER_n&=\int \left(\begin{matrix}1\\ \frac{X_1-x}h \end{matrix} \right) \left[m(X_1)-m(x)-\nabla m^T(x)(X_1-x)\right] h^{-p} K\left(\frac{X_1-x}{h}\right)f(X_1)dX_1\\
            &=\int\left(\begin{matrix}1\\ u\end{matrix}\right) \left[m(x+uh)-m(x)-h\nabla m^T(x)u\right] K(u)f(x+uh)du\\
            &=\int\left(\begin{matrix}1\\ u\end{matrix}\right)\left[\frac{h^2}2u^T \nabla^2 m(x) u+ \frac{h^3}{3!} D_m^3(x,u)\right]K(u)f(x+uh)du,\\
        \end{split}
    \end{equation*}

    Do Taylor expansion to $f(x+uh)$ and we conclude that 
    \begin{equation*}
        \begin{split}
            \int u^T\nabla^2m(x)uK(u)f(x+uh)du=&f(x)\mu_2\text{Tr}(\nabla^2m(x))+O(h^2),\\
            \int D_m^3(x,u)K(u)f(x+uh)du =& O(h),\\
            (\mathbb ER_n)_1=& \frac 12h^2f(x)\mu_2\text{Tr}(\nabla^2m(x))+O(h^4)\\
            \int u[u^T\nabla ^2m(x)u]K(u)f(x+uh)=&h\int u[u^T\nabla ^2m(x)u]\nabla f(x)^T uK(u)du + O(h^3),\\
            \int uD_m^3(x,u)K(u)f(x+uh)du=&f(x)\int uD_m^3(x,u)K(u)du +O(h^3),\\
            (\mathbb ER_n)_{2:p+1}=& \frac 12h^3\int u[u^T\nabla ^2m(x)u]\nabla f(x)^T uK(u)du\\
            &+\frac1{3!}f(x)h^3\int uD_m^3(x,u)K(u)du+O(h^5).\\
        \end{split}
    \end{equation*}
    
    Denote that $b(m)=\int uD_m^3(x,u)K(u)du$, $b_1(m)=\int u[u^T\nabla^2 m(x)u]\nabla f(x)^TuK(u)du-\mu_2^2\nabla f(x)\text{Tr}(\nabla^2m(x))$, and combine the above conclusions:
    \begin{equation*}
        \begin{split}
            &\mathbb E(\widehat{\theta}_{(n)} -\theta|X_1,\dots ,X_n) \\=& \text{diag}\{1,h^{-1}I_p\} S_n^{-1}R_n \\
            =& \text{diag}\{1,h^{-1}I_p\} (A(h)^{-1}+ O(\{nh^p\}^{-1/2}))R_n \\
            =& \left(\begin{matrix} 1\\ &h^{-1}I_p\end{matrix}\right)\left\{\left( \begin{matrix} \frac1{f(x)} & -\frac h{f^2(x)}\nabla f(x)^T\\ -\frac h{f^2(x)}\nabla f(x) & \frac1{\mu_2f(x)}I_p \end{matrix} \right)+ O(h^2)+ O(\{nh^p\}^{-1/2})\right\}\\
            &\cdot\left(\begin{matrix}
                \frac12h^2f(x)\mu_2\text{Tr}(\nabla^2m(x))+O(h^4)\\
                \frac12h^3\int u[u^T\nabla m(x) u]\nabla f(x)^T uK(u)du +\frac1{3!}f(x)h^3\int uD_m(x,u)K(u) du +O(h^5)
            \end{matrix}\right)\\
            =&\left(\begin{matrix}
                \frac12h^2f(x)\mu_2\text{Tr}(\nabla^2m(x))+ O(h^4)+ O(n^{-1/2} h^{2-p/2})\\
                \frac{h^2}{2\mu_2f(x)}b_1(m) +\frac{h^2}{3!\mu_2}b(m) +O(h^4) +O(n^{-1/2} n^{2-p/2})
            \end{matrix}\right)\\
            =&h^2\left(\begin{matrix}
                \frac12f(x)\mu_2\text{Tr}(\nabla^2m(x))\\
                \frac{1}{2\mu_2f(x)}b_1(m) +\frac{1}{3!\mu_2}b(m)
            \end{matrix}\right)+O(h^4)+O(n^{-1/2}h^{2-p/2}).
    \end{split}
    \end{equation*}
    
    For the covariance of $\widehat{\theta}_{(n)}$ , we have 
    \begin{equation*}
        \begin{split}
            \widehat{\theta}_{(n)}-\theta^* &= (\mathbf X\mathbf W\mathbf X^T) ^{-1} \mathbf{X}\mathbf{W}(\mathbf M+ \varepsilon-\theta^{*T}\mathbf X)\\
            &=\left(\begin{matrix}1&\\ &h^{-1} I_p\end{matrix}\right) S_n^{-1} R_n + \left(\begin{matrix}1&\\ &h^{-1} I_p\end{matrix}\right) S_n^{-1} Z_n,
        \end{split}
    \end{equation*}
    which implies that 
    \begin{equation*}
        \left(\begin{matrix}1&\\ &h^{} I_p\end{matrix}\right)\left(\theta_{(n)}^* - \theta^*-\left(\begin{matrix}1&\\ &h^{-1} I_p\end{matrix}\right) S_n^{-1}R_n\right)=S_n^{-1}Z_n,
    \end{equation*}
    where 
    \begin{equation*}
        Z_n = \frac1n \sum_{i=1}^n\left(\begin{matrix}1\\\frac{X_i-x}h\end{matrix}\right)K\left(\frac{X_i-x}h\right)\varepsilon_i.
    \end{equation*}
    By CLT, we simply get that 
    \begin{equation*}
        Z_n\to_d N\left(0, \frac{\sigma^2}{nh^p} f(x)\left(\begin{matrix}J_0&\\ &J_2I_p\end{matrix}\right)\right).
    \end{equation*}
    
    Thus, 
    \begin{equation*}
        \begin{split}
            S_n^{-1}Z_n &\to_d N\left(0,\frac{\sigma^2}{nh^p} f(x)S_n^{-1} \left( \begin{matrix} J_0 &\\ &J_2I_p \end{matrix} \right) S_n^{-1}\right)\\
            &=N\left(0,\frac{\sigma^2}{nh^pf(x)} \left\{\left(\begin{matrix} J_0&\\ &\frac{J_2}{\mu_2^2h^2}I_p \end{matrix} \right)+O(h^{2}) + O(n^{-1/2}h^{-p/2})\right\}\right)\\
            \text{Cov}(\widehat{\theta}_{(n)}|X_1,\dots,X_n) &= \frac{\sigma^2}{nh^pf(x)}\left\{\left(\begin{matrix}J_0&\\ &\frac{J_2}{\mu_2^2h^2}I_p\end{matrix}\right)+O(h^{2}) + O(n^{-1/2}h^{-p/2})\right\}.
        \end{split}
    \end{equation*}
\end{proof}

\subsection{Proof of Theorem \ref{thm:estimation}} \label{appsubsec:thm_estimation}
\begin{proof}
    Decompose expected error of the estimation $\widehat{\theta}_{{(N)}}$ as the following format
    \begin{equation*}
        \begin{split}
            \mathbb E(\hat \theta_{(N)}-\theta^*|\mathcal L,\mathcal U) & = \mathbb E(\hat \theta_{(N)}|\mathcal L,\mathcal U)-\theta^*\\
            &=(\widetilde{\mathbf X}\widetilde{\mathbf W}\widetilde{\mathbf X}^T)^{-1}\widetilde{\mathbf X}\widetilde{\mathbf W}(\widetilde{\mathbf M}-\widetilde{\mathbf X}^T\theta^*) - (\widetilde{\mathbf X}\widetilde{\mathbf W}\widetilde{\mathbf X}^T)^{-1}\widetilde{\mathbf X}\widetilde{\mathbf W}\widetilde{\mathbf r}+ (\mathbf X{\mathbf W}{\mathbf X}^T)^{-1}\mathbf X{\mathbf W}{\mathbf r}.
        \end{split}
    \end{equation*}
    
    We have $(\widetilde{\mathbf X}\widetilde{\mathbf W}\widetilde{\mathbf X}^T)^{-1}\widetilde{\mathbf X}\widetilde{\mathbf W}(\widetilde{\mathbf M}-\widetilde{\mathbf X}^T\theta^*)\to N(B_L(x,h)+O(h^4)+O(N^{-1/2}h^{2-p/2}), \sigma^2O(N^{-1}h^p))$.
    Then we need to derive the corresponding distribution of $(\mathbf X{\mathbf W}{\mathbf X}^T)^{-1}\mathbf X{\mathbf W}{\mathbf r} - (\widetilde{\mathbf X}\widetilde{\mathbf W}\widetilde{\mathbf X}^T)^{-1}\widetilde{\mathbf X}\widetilde{\mathbf W}\widetilde{\mathbf r}$.
    
    \begin{equation*}
        \begin{split}
            &(\mathbf X{\mathbf W}{\mathbf X}^T)^{-1}\mathbf X{\mathbf W}{\mathbf r} - (\widetilde{\mathbf X}\widetilde{\mathbf W}\widetilde{\mathbf X}^T)^{-1}\widetilde{\mathbf X}\widetilde{\mathbf W}\widetilde{\mathbf r}\\
            =&\left(\begin{matrix}1&\\&h^{-1}I_p\end{matrix} \right) S_n^{-1} R_n^{(r)} - \left( \begin{matrix}1&\\&h^{-1}I_p\end{matrix} \right) \widetilde S_N^{-1}\widetilde R_N^{(r)}\\
            =&\left\{ \left( \begin{matrix}1\\&h^{-1}I_p \end{matrix}\right) \left\{\left( \begin{matrix}\frac1{f(x)} & -\frac h{f^2(x)}\nabla f(x)^T\\-\frac h{f^2(x)}\nabla f(x) & \frac1{\mu_2f(x)}I_p \end{matrix}\right) +O(h^2) +O(\{nh^p\}^{-1/2})\right\} R_n^{(r)} \right\}\\
            &-\left\{ \left(\begin{matrix}1\\&h^{-1}I_p\end{matrix}\right) \left\{\left(\begin{matrix}\frac1{f(x)} & -\frac h{f^2(x)} \nabla f(x)^T\\-\frac h{f^2(x)}\nabla f(x) & \frac1{\mu_2f(x)}I_p \end{matrix}\right) +O(h^2)+O(\{Nh^p\}^{-1/2})\right\} \widetilde R_N^{(r)}\right\}.\\
        \end{split}
    \end{equation*}
    
    Since the expectation of $R_n^{(r)}$ and $R_N^{(r)}$ are the same due to their definition
    \begin{equation*}
        \begin{split}
            R_n^{(r)} &= \frac1n \sum_{i=1}^n h^{-p}\left(\begin{matrix} 1\\\frac{X_i-x}h\end{matrix}\right) K\left(\frac{X_i-x}h\right) [F(X_i)- m(X_i)],\\
            \widetilde R_N^{(r)} &= \frac1N \sum_{i=1}^N h^{-p}\left( \begin{matrix} 1\\\frac{\widetilde X_i-x}h\end{matrix}\right) K\left(\frac{\widetilde X_i-x}h\right) [F(\widetilde X_i)- m(\widetilde X_i)],\\
            \mathbb ER_n^{(r)} &= \int \left(\begin{matrix} 1\\u\end{matrix} \right) K(u)r(x+uh)f(x+uh)du = \mathbb E\widetilde R_N^{(r)},
        \end{split}
    \end{equation*}
    we derive the distribution of their difference as 
    \begin{equation*}
        \sqrt{\frac{Nn}{N+n}h^p}(R_n^{(r)}-\widetilde R_N^{(r)}) \to_d N\left(0, \int \left( \begin{matrix}1&u^T\\u&uu^T\end{matrix} \right) K^2(u) r(x+uh) f(x+uh) du\right).
    \end{equation*}
    
    Consequently, the distribution of the rectifier $\hat \Delta$ can be derived as
    \begin{equation*}
        \begin{split}
            &(\mathbf X{\mathbf W}{\mathbf X}^T)^{-1}\mathbf X{\mathbf W}{\mathbf r} - (\widetilde{\mathbf X}\widetilde{\mathbf W}\widetilde{\mathbf X}^T)^{-1}\widetilde{\mathbf X}\widetilde{\mathbf W}\widetilde{\mathbf r}\\
            =&\ \left\{ \left(\begin{matrix}1\\&h^{-1}I_p\end{matrix}\right) \left\{\left(\begin{matrix}
                \frac1{f(x)} & -\frac h{f^2(x)}\nabla f(x)^T\\
                -\frac h{f^2(x)}\nabla f(x) & \frac1{\mu_2f(x)}I_p 
            \end{matrix}\right)+O(h^2)+O(\{nh^p\}^{-1/2}) \right\}R_n^{(r)}\right\}\\
            &-\left\{ \left(\begin{matrix}1\\&h^{-1}I_p\end{matrix} \right) \left\{\left(\begin{matrix}
                \frac1{f(x)} & -\frac h{f^2(x)}\nabla f(x)^T\\
                -\frac h{f^2(x)}\nabla f(x) & \frac1{\mu_2f(x)}I_p 
            \end{matrix}\right)+O(h^2)+O(\{Nh^p\}^{-1/2})\right\}\widetilde R_N^{(r)}\right\}\\
            =&\ \left(\begin{matrix}1\\&h^{-1}I_p\end{matrix}\right)\left\{\frac1{f(x)} \left(\begin{matrix}2 & -\frac h{f (x)}\nabla f(x)^T\\-\frac h{f(x)}\nabla f(x) & \frac1{\mu_2}I_p \end{matrix}\right)+O(h^2)\right\}\left(R_n^{(r)}-\widetilde R_N^{(r)}\right)\\
            &\ +\left(\begin{matrix}1\\&h^{-1}I_p\end{matrix}\right) R_n^{(r)} O(\{nh^p\}^{-1/2}) \\
            =&\ \left(\begin{matrix}1\\&h^{-1}I_p\end{matrix}\right)
            \left\{\frac1{f(x)}\left(\begin{matrix}2 & -\frac h{f (x)}\nabla f(x)^T\\-\frac h{f(x)}\nabla f(x) & \frac1{\mu_2}I_p \end{matrix}\right)+O(h^2)\right\}o_p(\{nh^p\}^{-1/2})\\
            &\ +\left(\begin{matrix}1\\&h^{-1}I_p\end{matrix}\right) (\mathbb ER_n^{(r)}+o_p(\{nh^p\}^{-1/2})) O(\{nh^p\}^{-1/2})\\
            =&\ \left(\begin{matrix}1\\&h^{-1}I_p\end{matrix}\right) (\mathbb ER_n^{(r)}+o_p(1)) O(\{nh^p\}^{-1/2}).
        \end{split}
    \end{equation*}
    
    As result, we have the expected error of $\widehat{\theta}_{(N)}$
    \begin{equation*}
        \begin{split}
            \mathbb E(\hat \theta_{(N)}-\theta^*|\mathcal L,\mathcal U) &= B_L(x,h) + \left(\begin{matrix}1\\&h^{-1}I_p\end{matrix}\right)(\mathbb ER_n^{(r)}+o_p(1)) O(\{nh^p\}^{-1/2})+O(h^4)\\
            &=O(h^2)+O(n^{-1/2}h^{-1-p/2})\to_d 0.
        \end{split}
    \end{equation*}
\end{proof}
    
\subsection{Proof of Theorem \ref{thm:res_var}}\label{appsubsec:thm_res_var}
\begin{proof}
    We derive the variance of $\widehat{\theta}^{\text{con}}$ and $\widehat{\theta}^{\text{PP}}$ under the expectation of $\varepsilon$, $\mathcal L$ and $\mathcal U$.
    
    \begin{equation*}
        \begin{split}
            \text{Cov}(\widehat{\theta}^{\text{con}}) =& \text{Cov}\left((\mathbf{X}\mathbf{W}\mathbf{X}^T)^{-1}\mathbf{X}\mathbf{W}(\mathbf{M}+\varepsilon)\right) \\
            =& \text{Cov}\left(\left( \begin{matrix}1&\\&h^{-1}I_p\end{matrix} \right) \left(\frac1n\sum_{i=1}^nh^{-p}\left(\begin{matrix}1\\ \frac{X_i-x}h\end{matrix}\right)\left( \begin{matrix} 1&\frac{X_i-x}h\end{matrix} \right) K\left(\frac{X_i-x}h \right) \right)^{-1} \right.\\
            &\cdot\left.\left(\frac1n\sum_{i=1}^nh^{-p} \left( \begin{matrix} 1\\\frac{X_i-x}h \end{matrix} \right) K\left( \frac{X_i-x}h\right) \left(m(X_i) +\varepsilon_i\right) \right)\right) \\
            =&\text{Cov}\left(\left(\begin{matrix} 1&\\&h^{-1}I_p\end{matrix} \right)(A(h)^{-1}+O(h^2))\left(\frac1n \sum_{i=1}^n h^{-p} \left( \begin{matrix} 1\\\frac{X_i-x}h \end{matrix}\right) K\left(\frac{X_i-x}h \right) \left(m(X_i) +\varepsilon_i\right) \right) \right) \\
            =& n^{-1}A^{-1}(h)\text{Cov}\left(h^{-p} \left( \begin{matrix} 1&\\&h^{-1}I_p \end{matrix} \right) X_1^+K_1(M(X_1)+\varepsilon_1) \right)A^{-1}(h) + O(h^2). \\
        \end{split}
    \end{equation*}
    where $A(h)^{-1}$ is defined in Appendix \ref{appsec:proof} and $X_1^+ = \left(\begin{matrix}1\\\frac{X_1-x}{h}\end{matrix}\right)$.
    Actually we have $\text{Cov}(\widehat{\theta}^{\text{con}})\sim\Omega(1)n^{-1}h^{-2p}I$.
    
    Decompose the middle term $\text{Cov}\left(M_hX_1^+K_1(M(X_1)+\varepsilon_1)\right)$ locally where $M_h=h^{-p}\left(\begin{matrix}1&\\&h^{-1}I_p\end{matrix}\right)$, 
    
    \begin{equation*}
        \begin{split}
            &\text{Cov}\left(M_hX_1^+K_1(M(X_1)+\varepsilon_1)\right)\\
            =&\mathbb E\left(M_hX_1^+X_1^{+T}K_1^2(M(X_1)+\varepsilon_1)^2M_h\right) - \mathbb E\left(M_hX_1^+K_1(M(X_1)+\varepsilon_1)\right)\mathbb E\left(M_hX_1^{+T}K_1(M(X_1)+\varepsilon_1)\right) \\ 
            =&\mathbb E\left(M_hX_1^+X_1^{+T}K_1^2M(X_1)^2M_h\right)+2\mathbb E\left(M_hX_1^+X_1^{+T}K_1^2M(X_1)\varepsilon_1M_h\right) + \mathbb E\left(M_hX_1^+X_1^{+T}K_1^2\varepsilon_1^2M_h\right)\\
            &- \mathbb E\left(M_hX_1^+K_1M(X_1)\right)\mathbb E\left(M_hX_1^{+T}K_1M(X_1)\right) - \mathbb E\left(M_hX_1^+K_1M(X_1)\right)\mathbb E\left(M_hX_1^{+T}K_1\varepsilon_1\right)\\
            &- \mathbb E\left(M_hX_1^+K_1\varepsilon_1\right)\mathbb E\left(M_hX_1^{+T}K_1M(X_1)\right) - \mathbb E\left(M_hX_1^+K_1\varepsilon_1\right)\mathbb E\left(M_hX_1^{+T}K_1\varepsilon_1\right)\\
            =&\mathbb E\left(M_hX_1^+X_1^{+T}K_1^2M(X_1)^2M_h\right)+\mathbb E\left(M_hX_1^+X_1^{+T}K_1^2\varepsilon_1^2M_h\right)- \mathbb E\left(M_hX_1^+K_1M(X_1)\right)\mathbb E\left(M_hX_1^{+T}K_1M(X_1)\right) \\
            =&\text{Cov}\left(M_hX_1^+K_1M(X_1)\right)+\sigma^2\mathbb E\left(M_hX_1^+X_1^{+T}K_1^2M_h\right).\\
        \end{split}
    \end{equation*}
    
    Do the same procedure to $\widehat{\theta}^{\text{PP}}$,
    \begin{equation*}
        \begin{split}
            \text{Cov}(\widehat{\theta}^{\text{PP}}) =& \text{Cov}\left((\mathbf{X}\mathbf{W}\mathbf{X}^T)^{-1}\mathbf{X}\mathbf{W}(\mathbf{F}-\mathbf{M}-\varepsilon)\right) + \text{Cov}\left((\mathbf{\widetilde X}\mathbf{\widetilde W}\mathbf{\widetilde X}^T)^{-1}\mathbf{\widetilde X}\mathbf{\widetilde W}(\mathbf{\widetilde F})\right) \\
            =& n^{-1}A^{-1}(h)\text{Cov}\left(M_hX_1^+K_1(F(X_1)-m(X_1))\right)A^{-1}(h) \\
            &+n^{-1}A^{-1}(h)\sigma^2\mathbb E\left(M_hX_1^+X_1^{+T}K_1^2M_h\right)A^{-1}(h)\\
            &+ N^{-1}A^{-1}(h)\text{Cov}\left(M_hX_1^+K_1F(X_1)\right)A^{-1}(h) +O(h^2).
        \end{split}
    \end{equation*}
    
    Then, we do subtraction to these two covariance matrix and gain
    \begin{equation*}
        \begin{split}
            A(h)[\text{Cov}(\widehat{\theta}^{\text{con}}) - \text{Cov}(\widehat{\theta}^{\text{PP}})]A(h) =&  n^{-1}[\text{Cov}\left(M_hX_1^+K_1m(X_1)\right)-\text{Cov}\left(M_hX_1^+K_1(F(X_1)-m(X_1))\right)]\\
            & -N^{-1}\text{Cov}\left(M_hX_1^+K_1F(X_1)\right) + O(h^2).
        \end{split}
    \end{equation*}
    
    
    Given the formula each term as:
    \begin{equation*}
        \begin{split}
            &\text{Cov}\left(M_hX_1^+K_1m(X_1)\right) \\
            = &M_h^2\left(\begin{matrix}
                m^2(x)f(x)J_0 + O(h^2)& h[m^2(x)\nabla f(x)^T+2m(x)f(x)\nabla m(x)^T]+ O(h^3) \\
                h[m^2(x)\nabla f(x)+2m(x)f(x)\nabla m(x)] + O(h^3)& m^2(x)f(x)J_2I_p + O(h^2)
            \end{matrix}\right)\\
            = &\frac{m^2(x)f(x)}{h^{-2p}}\left(\begin{matrix}J_0 & \\ & J_2h^{-2}I_p\end{matrix}\right)+O(h^{-2p-2}),
        \end{split}
    \end{equation*}
    and consequently, Under $F\in C^2(U)$, we have
    \begin{equation*}
        \begin{split}
            &\text{Cov}\left(M_hX_1^+K_1(F(X_1)-m(X_1))\right)         =   \frac{[F(x)-m(x)]^2f(x)}{h^{-2p}} \left(\begin{matrix}J_0 & \\ & J_2h^{-2}I_p \end{matrix} \right)+O(h^{-2p-2}).
        \end{split}
    \end{equation*}
    
    Thus, $\text{Cov}\left(M_hX_1^+K_1(F(X_1)-m(X_1))\right)\ll \text{Cov}\left(M_hX_1^+K_1m(X_1)\right)$ appears to hold if $[F(x)-m(x)]^2\ll m^2(x)$ in expectation, which is promised by the superiority of the predictor $F$ with respect to $m(x)$.

    Thus, we have 

    \begin{equation*}
        \begin{split}
            &\text{Cov}(M_hX_1^+K_1m(X_1))-\text{Cov}(M_hX_1^+K_1(F(X_1)-m(X_1)))\\
            =&\frac{[m^2(x)-(m(x)-F(x))^2]f(x)}{h^{-2p}}\left(\begin{matrix}J_0 & \\ & J_2h^{-2}I_p \end{matrix} \right)+O(h^{-2p-2})\\
            &\text{Cov}(M_hX_1^+K_1F(X_1))\\
            =&\frac{F(x)^2f(x)}{h^{-2p}}\left(\begin{matrix}J_0 & \\ & J_2h^{-2}I_p \end{matrix} \right)+O(h^{-2p-2})\\
        \end{split}
    \end{equation*}
    
    \begin{equation*}
        \begin{split}
            &\text{Cov}(\widehat{\theta}^{\text{con}}) - \text{Cov}(\widehat{\theta}^{\text{PP}}) \\
            =&[\frac{m^2(x)-(m(x)-F(x))^2}{n}-\frac{F^2(x)}{N}]\frac{f(x)}{h^{-2p}}A^{-1}(h)\left(\begin{matrix}J_0 & \\ & J_2h^{-2}I_p \end{matrix} \right)A^{-1}(h) + O(h^{-2p-2})\\
            =&[\frac{m^2(x)-(m(x)-F(x))^2}{n}-\frac{F^2(x)}{N}]\frac{1}{h^{-2p}f(x)} (h)\left(\begin{matrix}J_0 & \\ & \frac{J_2}{h^2\mu_2}I_p \end{matrix} \right) (h) + O(h^{-2p-2})\\
            \succ & c_0 n^{-1}h^{-2p} I
        \end{split}
    \end{equation*}
        where $c_0=\Omega(1)< [{m^2(x)-(m(x)-F(x))^2}-{F^2(x)}\frac n{N}]\frac{\min\{J_0, h^{-2}\mu_2^{-2}J_2\}}{f(x)}$. 
        Since $N\gg n$ and $m^2(x) \gg [m(x)-F(x)]^2$, we just take $N>\gamma n$ and $m^2(x) \approx F^2(x) > \gamma[m(x)-F(x)]^2$, one of lower bound is
    \begin{equation*}
        \begin{split}
            \frac{m^2(x)-(m(x)-F(x))^2}{n}-\frac{F^2(x)}{N}>&\frac{(1-1/\gamma)m^2(x)}{n}-\frac{F^2(x)}{N}\frac{\min\{J_0, h^{-2}\mu_2^{-2}J_2\}}{h^{-2p}f(x)}\\
            >&\frac{(\gamma-1)m^2(x)-F^2(x)}{N}\frac{\min\{J_0, h^{-2}\mu_2^{-2}J_2\}}{h^{-2p}f(x)}\\
            \approx & \frac{(\gamma-2)m^2(x)}{N}\frac{\min\{J_0, h^{-2}\mu_2^{-2}J_2\}}{h^{-2p}f(x)}\\
            >&\frac{m^2(x)}{2n}\frac{\min\{J_0, h^{-2}\mu_2^{-2}J_2\}}{h^{-2p}f(x)}:=c_0n^{-1}h^{-2p}.\\
        \end{split}
    \end{equation*}
        
\end{proof}
    
\subsection{Proof of Theorem \ref{thm:CIdeduction}}
\begin{proof}
    The conclusion of $|\mathcal C_{1,\alpha}^{\text{PP}}|<|\mathcal C_{1,\alpha}^{\text{con}}|$ can be derived through the conclusion of Theorem \ref{thm:res_var} since $\sigma_{1,1}^{\text{PP}}<\sigma_{1,1}^{\text{con}}$.
    
    For the volume of $\mathcal C_{2:p+1,\alpha}^{\text{PP}}$ and $\mathcal C_{2:p+1,\alpha}^{\text{con}}$, define $A=\text{Cov}(\widehat{\nabla m(x)}^{\text{con}})$ and $B=\text{Cov}(\widehat{\nabla m(x)}^{\text{PP}})$ so that $\mathcal C_{2:p+1,\alpha}^{\text{PP}} = \{u+\widehat\theta^{\text{PP}}:u^TB^{-1}u\leq \chi^2_p(1-\alpha)\}$ and $\mathcal C_{2:p+1,\alpha}^{\text{con}} = \{u+\widehat\theta^{\text{con}}:u^TA^{-1}u\leq \chi^2_p(1-\alpha)\}$.
    To compare the volume of such two sets, it's equivalent to compare $\{u^TA^{-1}u\leq c\}$ and $\{u^TB^{-1}u\leq c\}$. 
    
    Since $A\succ B$ through the conclusion of Theorem \ref{thm:res_var}, we have $B^{-1}\succ A^{-1}$.
    Consequently, for any vector $u$, $B^{-1}\succ A^{-1}$ implies $u^TB^{-1}u>u^TA^{-1}u$, which means that $u+\widehat{\nabla m(x)}^{\text{con}}\in \mathcal C_{2:p+1,\alpha}^{\text{con}}\iff u^TB^{-1}u\leq c\Longrightarrow u+\widehat{\nabla m(x)}^{\text{PP}}\in \mathcal C_{2:p+1,\alpha}^{\text{PP}}$.
    Thus, we have the volume of $\mathcal C_{2:p+1,\alpha}^{\text{PP}}$ is smaller than $\mathcal C_{2:p+1,\alpha}^{\text{con}}$.
        
\end{proof}
    
\subsection{Proof of Theorem \ref{thm:coverageprob}}
    
\begin{proof}
    
    For the coverage probability of confidence interval of $\hat{m}(x)^{\text{PP}}=\theta_1^{\text{PP}}$, i.e. $\mathcal C_{1,\alpha}$ in Equation \eqref{equ:CIone}, we have
    \begin{equation*}
        \begin{split}
            &\mathbb P\left\{m(x)\in[\hat{m}(x)^{\text{PP}} - z_{1-\alpha/2}\sigma_{1,1}, \hat{m}(x)^{\text{PP}} + z_{1-\alpha/2}\sigma_{1,1}]\right\}\\
            =&\int_{{m}(x) - z_{1-\alpha/2}\sigma_{1,1}}^{{m}(x) + z_{1-\alpha/2}\sigma_{1,1}} \frac1{\sqrt{2\pi\sigma_{1,1}^2}}\exp\left\{-\frac{[t-m(x)-\frac12h^2B_1(x)-R(x,h)]^2}{2\sigma_{1,1}^2}\right\}dt\\
            =&\int_{-z_{1-\alpha/2}}^{z_{1-\alpha/2}} \frac1{\sqrt{2\pi}}\exp\left\{-\frac{u^2}2\right\}\exp\left\{-\frac{h^4}{8\sigma_{1,1}^2}B_1(x)-\frac1{2\sigma_{1,1}^2}R(x,h)^2+u\frac{h^2B_1(x)+2R(x,h)}{4\sigma_{1,1}^2}\right.\\
            &\left.-\frac{h^2}{2\sigma_{1,1}^2}B_1(x)R(x,h)\right\}du\\
            =&\int_{-z_{1-\alpha/2}}^{z_{1-\alpha/2}} \frac1{\sqrt{2\pi}}\exp\left\{-\frac{u^2}2\right\}\left(1-\frac{h^4}{8\sigma_{1,1}^2}B_1(x)+u\frac{h^2B_1(x)+2R(x,h)}{4\sigma_{1,1}^2}+O(h^6)+O(n^{-1/2}h^{2-p/2})\right)du\\
            =&(1-\alpha)\left(1-\frac{h^4}{8\sigma_{1,1}^2}B_1(x)+O(h^6)+O(n^{-1/2}h^{2-p/2})\right).
        \end{split}
    \end{equation*}
    where $B_1(x)=f(x)\mu_2\text{Tr}(\nabla^2m(x))$.
    
    For multivariable gradient estimation, we suppose the actually distribution is 
    $$
    \hat \theta -\theta^* - B_g \sim N(0,\Sigma).\\
    $$
    We build the confidence set of $\theta^*$ as 
    $$
    \mathcal C_{2:p+1} = \{\theta: (\theta-\widehat{\theta})^T\Sigma^{-1}(\theta-\widehat{\theta})\leq \chi^2_p(1-\alpha)\},
    $$
    and compute that 
    $$
    \mathbb P((\theta^*-\widehat{\theta})^T\Sigma^{-1}(\theta^*-\widehat{\theta})\leq \chi^2_p(1-\alpha)),
    $$
    where $\chi^2_p(1-\alpha)$ satisfies that $P(u^Tu\leq \chi_p^2(1-\alpha))=1-\alpha$ where $u\sim N(0,I_p)$. More specifically, 
    $$
    1-\alpha = \int_0^{\chi_p^2(1-\alpha)} \frac{x^{p/2-1}e^{-x/2}}{2^{p/2}\Gamma (p/2)}dx .
    $$
    However, according to the ground truth about $\widehat{\theta}$, we have $\mathbb P((\theta^*-\widehat{\theta}+B_g)^T\Sigma^{-1}(\theta^*-\widehat{\theta}+B_g)\leq \chi_p^2(1-\alpha))$
    \begin{equation}
        \begin{split}
            &\mathbb P\left((\theta^*-\widehat{\theta})^T\Sigma^{-1}(\theta^*-\widehat{\theta})\leq \chi^2_p(1-\alpha)|\widehat{\theta}\sim N(\theta^*+B_g, \Sigma)\right) \\
            =& \mathbb P\left(u^T\Sigma^{-1}u\leq \chi_p^2(1-\alpha)|u\sim N(B_g, \Sigma)\right)\\
            =& \mathbb P\left(u^Tu\leq \chi_p^2(1-\alpha )\big|u\sim N(\Sigma^{-1/2}B_g, I_p)\right)\\
            =& \mathbb P\left(\sum_{i=1}^p u^2_i \leq \chi^2_p(1-\alpha)\bigg| u_i\sim N((\Sigma^{-1/2}B_g)_i, 1)\right).
        \end{split}
    \end{equation}
    Define $\Sigma^{-1/2}B_g = \{b_i, i\in[p]\}$.  The probability density function can be calculated as:
    \begin{equation*}
        \begin{split} 
            \mathbb P(u_i^2\leq y) =& \mathbb P(-\sqrt y\leq u_i\leq \sqrt y)\\
            =&\int_{-\sqrt y}^{\sqrt y}\frac1{\sqrt{2\pi}}\exp\{-\frac12(x-b_i)^2\}dx,\\
            \text{p.d.f}(y) = & \frac1{\sqrt {2\pi}}\exp\{-\frac12(\sqrt y-b_i)^2\}(\sqrt y)'-\frac1{\sqrt {2\pi}}\exp\{-\frac12(-\sqrt y-b_i)^2\}(-\sqrt y)'\\
            = & \frac1{\sqrt {2\pi}}\exp\{-\frac12( y+b_i^2-2b_i\sqrt y)\}\cdot\frac1{2\sqrt y}+\frac1{\sqrt {2\pi}}\exp\{-\frac12(y+b_i^2+2b_i\sqrt y)\}\cdot\frac1{2\sqrt y}\\
            = & \frac1{2\sqrt {2\pi y}}\exp\{-\frac12( y+b_i^2)\}(\exp\{-\sqrt yb_i\}+\exp\{\sqrt yb_i\}).\\
        \end{split}    
    \end{equation*}
    Notice that for conventional treatment, the covariance matrix can be described as
    \begin{equation*}
        \begin{split}
            \Sigma =& \frac{\sigma^2J_2}{nh^{p+2}f(x)\mu_2^2}I_p+O(n^{-1}h^{2-p})+O(n^{-3/2}h^{-3p/2}),\\
            \Sigma^{-1/2} =& \frac{\sigma\sqrt{J_2}}{\sqrt{nh^{p+2}f(x)} \mu_2}I_p +O(n^{-1/2}h^{3-p/2})+O(n^{-1}h^{1-p}),\\
            B_g =&\frac{h^2}{2\mu_2f(x)}b_1(m)+\frac{h^2}{6\mu_2}b(m) + O(h^4)+O(n^{-1/2}h^{-p/2}).
        \end{split}
    \end{equation*}
    Thus, 
    $$
    \{b_i\} = \Sigma^{-1/2}B_g = \frac{\sigma\sqrt{J_2h^2}}{\sqrt{nh^{p}f(x)}\mu_2^2}\left(\frac1{2f(x)}b_1(m)+\frac16b(m)\right)+O\left(n^{-1/2}h^{3-p/2}\right)+O\left(n^{-1}h^{-1-p}\right).
    $$
    Denote $C_b =\frac{\sigma\sqrt J_2}{\sqrt {f(x)}\mu_2^2}(\frac1{2f(x)}b_1(m)+\frac16 b(m))$, then $\{b_i\}=(C_b+O(h^2)+O(n^{-1/2}h^{-p/2}))n^{-1/2}h^{1-p/2}$.  Thus, the p.d.f of $\mathbb P(u_i^2\leq y)$ is 
    \begin{equation*}
        \begin{split}
            \text{p.d.f}(y) = &\frac1{2\sqrt {2\pi y}}\exp\{-\frac12( y+b_i^2)\}(\exp\{-\sqrt yb_i\}+\exp\{\sqrt yb_i\})\\
            =&\frac {y^{-1/2}\exp\{-\frac12 y\}}{\sqrt {2\pi}}\frac{1+\exp\{2\sqrt yb_i\}}{2\exp\{\sqrt yb_i\}}\exp\{-\frac12b_i^2\}\\
            =&\frac {y^{-1/2}\exp\{-\frac12 y\}}{2^{1/2}\Gamma(1/2)}\left[1+(y-\frac12)b_i^2+O(b_i^3)\right].
        \end{split}
    \end{equation*}
    Then we check the additivity of this biased Gamma distribution of convolution.
    
    Notice that we only care about the condition that $y\leq \chi_p^2(1-\alpha)$ which is a limited condition, thus we have $(y-\frac12)b_i^2$ is $o(1)$ under $y\leq \chi_p^2(1-\alpha)$. Let $\widetilde h = n^{-1/2}h^{1-p/2}$ and then $b_i = O(1)\widetilde h$. Denote a biased Gamma distribution where
    $$
    BGamma(y,\lambda, \alpha, b^2) = \frac{\lambda^\alpha y^{\alpha-1}\exp\{-\lambda y\}}{\Gamma (\alpha)}[1+(y-\alpha)b^2+O(\widetilde h^3)],\quad y\leq \chi_p^2(1-\alpha).
    $$
    Actually, $\Omega(\widetilde h^2)$ contains $(y-\alpha)b_i^2$ and $O(b_i^3)$ while the latter term contains higher order of $y$ but was neglected because of the enough small $b_i$.
    
    Define $Y=Y_1+Y_2$, and the notion $*$ refers to the convolution operation, then
    \begin{equation*}
        \begin{split}
            f(y) =& f_{Y_1}(y_1)*f_{Y_2}(y_2)\\
            =&\int_0^y \frac{\lambda^{\alpha_1}t^{\alpha_1-1}e^{-\lambda t}}{\Gamma(\alpha_1)}\frac{\lambda^{\alpha_2}(y-t)^{\alpha_2-1}e^{-\lambda (y-t)}}{\Gamma(\alpha_2)}\\
            &\cdot [1+(t-\alpha_1)b_1^2+O(\widetilde h^3)][1+(y-t-\alpha_2)b_2^2+O(\widetilde h^3)]dt\\
            =&\int_0^y \frac{\lambda^{\alpha_1+\alpha_2}t^{\alpha_1-1}(y-t)^{\alpha_2-1}e^{-\lambda y}}{\Gamma(\alpha_1)\Gamma(\alpha_2)}[1-\alpha_1b_1^2-\alpha_2b_2^2+O(\widetilde h^3)]dt\\
            &+\int_0^y \frac{\lambda^{\alpha_1+\alpha_2}t^{\alpha_1}(y-t)^{\alpha_2-1}e^{-\lambda y}}{\Gamma(\alpha_1)\Gamma(\alpha_2)}b_1^2dt+\int_0^y \frac{\lambda^{\alpha_1+\alpha_2}t^{\alpha_1-1}(y-t)^{\alpha_2}e^{-\lambda y}}{\Gamma(\alpha_1)\Gamma(\alpha_2)}b_2^2dt\\
            =&\frac{\lambda^{\alpha_1+\alpha_2}e^{-\lambda y}y^{\alpha_1+\alpha_2-1}\Gamma(\alpha_1)\Gamma(\alpha_2)}{\Gamma(\alpha_1+\alpha_2)\Gamma(\alpha_1)\Gamma(\alpha_2)}(1-\alpha_1b_1^2-\alpha_2b_2^2+O(\widetilde h^3))\\
            &+\frac{\lambda^{\alpha_1+\alpha_2}e^{-\lambda y}y^{\alpha_1+\alpha_2}\Gamma(\alpha_1+1)\Gamma(\alpha_2)}{\Gamma(\alpha_1+\alpha_2+1)\Gamma(\alpha_1)\Gamma(\alpha_2)}b_1^2+\frac{\lambda^{\alpha_1+\alpha_2}e^{-\lambda y}y^{\alpha_1+\alpha_2}\Gamma(\alpha_1)\Gamma(\alpha_2+1)}{\Gamma(\alpha_1+\alpha_2+1)\Gamma(\alpha_1)\Gamma(\alpha_2)}b_2^2\\
            =&\frac{\lambda^{\alpha_1+\alpha_2}e^{-\lambda y}y^{\alpha_1+\alpha_2-1}}{\Gamma(\alpha_1+\alpha_2)}\left[1-\alpha_1b_1^2-\alpha_2b_2^2+y(\frac{\alpha_1b_1^2+\alpha_2 b_2^2}{\alpha_1+\alpha_2})+O(\widetilde h^3)\right]\\
            =&\frac{\lambda^{\alpha_1+\alpha_2}e^{-\lambda y}y^{\alpha_1+\alpha_2-1}}{\Gamma(\alpha_1+\alpha_2)}\left[1+(y-\alpha_1-\alpha_2)(\frac{\alpha_1b_1^2+\alpha_2 b_2^2}{\alpha_1+\alpha_2})+O(\widetilde h^3)\right],
        \end{split}
    \end{equation*}
    which reduces to $\text{BGamma}(y, \lambda, \alpha_1+\alpha_2, (\alpha_1b_1^2+\alpha_2b_2^2)/(\alpha_1+\alpha_2))$. 
    
    By adding $Y_3$, we have new $\alpha$ is $\alpha_1+\alpha_2+\alpha_3$ and the new $b^2$ is 
    $$
    \frac{\frac{\alpha_1b_1^2+\alpha_2b_2^2}{\alpha_1+\alpha_2} (\alpha_1+\alpha_2)+\alpha_3b_3^2}{(\alpha_1+\alpha_2)+\alpha_3} = \frac{\sum_{i=1}^3 \alpha_ib_i^2}{\sum_{i=1}^3 \alpha_i}.
    $$
    This property can be generalized to $p$ components, which have the distribution of
    $$
    f_{\sum Y_i}(y) = \text{BGamma}\left(y, \lambda, \sum_{i=1}^p\alpha_i, \frac{\sum_{i=1}^p\alpha_ib_i^2}{\sum_{i=1}^p\alpha_i}\right).
    $$
    By taking $\lambda = \alpha_i=\frac12$, we have the p.d.f of biased chi squared statistic is 
    \begin{equation}
        \text{p.d.f}(y) = \frac{e^{-y/2}y^{p/2-1}}{2^{p/2}\Gamma(p/2)}\left(1+\frac{(y-p/2)}{p}\sum_{i=1}^p b_i^2+O(\widetilde h^3)\right),\quad y\leq \chi_p^2(1-\alpha).
    \end{equation}
    Thus, the coverage probability is 
    \begin{equation}
        \begin{split}
            \mathbb P(u^Tu\leq \chi_i^2(\alpha)) =& \int_0^{\chi_p^2(1-\alpha)}\frac{e^{-y/2}y^{p/2-1}}{2^{p/2}\Gamma(p/2)}\left(1+\frac{(y-p/2)}{p}\sum_{i=1}^p b_i^2+O(\widetilde h^3)\right)dy\\
            =&(1-\alpha)\left(1-\frac12\sum_{i=1}^p b_i^2+O(\widetilde h^3)\right)+\frac1p\sum_{i=1}^p b_i^2\int_0^{\chi_p^2(1-\alpha)}\frac{e^{-y/2}y^{(2+p)/2-1}}{2^{p/2}\Gamma(p/2)}dy\\
            =&(1-\alpha)\left(1-\frac12\sum_{i=1}^p b_i^2+O(\widetilde h^3)\right)+\sum_{i=1}^p b_i^2\int_0^{\chi_p^2(1-\alpha)}\frac{e^{-y/2}y^{(2+p)/2-1}}{2^{(p+2)/2}\Gamma((p+2)/2)}dy\\
            = &(1-\alpha)\left(1+\left(\frac12-c_1\right)\sum_{i=1}^p b_i^2+O(\widetilde h^3)\right).
        \end{split}
    \end{equation}
    where $c_1 = \int_{\chi_p^2(1-\alpha)}^{\chi_{p+2}^2(1-\alpha)}\frac{e^{-y/2}y^{(2+p)/2-1}}{2^{(p+2)/2}\Gamma((p+2)/2)}dy$ is a given constant related to $p$.

    As for the proposed confidence interval 
    \begin{equation*}
        \mathcal C_{1,\alpha}^{\text{BC}} = \left[\widehat{m(x)}^{\text{PP}} -h^2 B_1(x)  - z_{1-\alpha/2} \cdot\text{S.E.}\left(\widehat{m(x)}^{\text{PP}}\right), \widehat{m(x)}^{\text{PP}} -h^2B_1(x) + z_{1-\alpha/2} \cdot \text{S.E.}\left(\widehat{m(x)}^{\text{PP}}\right)\right],
    \end{equation*}
    and confidence set
    \begin{equation*}
        \begin{split}
            \mathcal C_{2:p+1, \alpha}^{\text{BC}} = &\left\{\nabla m(x) \bigg| \left( \widehat{\nabla m(x)}^{\text{PP}}-\nabla m(x)-B_2(x)\right)^T\right.\\
            &\left.\cdot\text{Cov}\left(\widehat{\nabla m(x)}^{\text{PP}}\right)^{-1}\left(\widehat{\nabla m(x)}^{\text{PP}}-\nabla m(x)-B_2(x)\right)\leq \chi_{p}^2(1-\alpha)\right\}
        \end{split}
    \end{equation*}
    with bias correction, it's equivalent to set $B_1(x)$ and $B_2(x)$ to zero for estimators $\widehat{m(x)}+h^2B_1(x)$ and $\widehat{\nabla m(x)}+B_2(x)$, which can eliminate the largest order terms of the error directly.

\end{proof}

\subsection{Proof of Theorem \ref{thm:high_dimensional}}
\begin{proof}
    We take expectation of $\hat\Delta_{(n)}^\text{HD}$ with respect to $\mathbf{\widetilde X}$ and $\mathbf{X}$, respectively.
    \begin{equation*}
        \begin{split}
            \mathbb E\hat\Delta^\text{HD}(t) &= \left(1+\frac{tN}n\right)\mathbb E_{\mathbf{X}}\left\{\mathbb E_{\mathbf{\widetilde X}}\left\{
            (\mathbf{X}\mathbf{W}\mathbf{X}^T+t\mathbf{\widetilde X}\mathbf{\widetilde W}\mathbf{\widetilde X}^T)^{-1}\mathbf{X}\mathbf{W}(\mathbf{Y}_F-\mathbf{Y})
            \right\}\right\}\\
            &=\left(1+\frac{tN}n\right)\mathbb E_{\mathbf{X}}\left\{\mathbb E_{\mathbf{\widetilde X}}\left\{
            (\mathbf{X}\mathbf{W}\mathbf{X}^T+t\mathbf{\widetilde X}\mathbf{\widetilde W}\mathbf{\widetilde X}^T)^{-1}\right\}\mathbf{X}\mathbf{W}(\mathbf{Y}_F-\mathbf{Y})
            \right\}\\
            &=\left(1+\frac{tN}n\right)\mathbb E_{\mathbf{X}}\left\{ 
            (\mathbf{X}\mathbf{W}\mathbf{X}^T+t\mathbb E_{\mathbf{\widetilde X}}\left\{\mathbf{\widetilde X}\mathbf{\widetilde W}\mathbf{\widetilde X}^T\right\})^{-1}\mathbf{X}\mathbf{W}(\mathbf{Y}_F-\mathbf{Y})
            \right\}\\
            &=\left(1+\frac{tN}n\right)\mathbb E_{\mathbf{X}}\left\{ 
            (\mathbf{X}\mathbf{W}\mathbf{X}^T+tN \mathbb EK_1X_1^+X_1^{+T})^{-1}\mathbf{X}\mathbf{W}(\mathbf{Y}_F-\mathbf{Y})
            \right\}\\
            &=\left(1+\frac{tN}n\right) 
            \left(\mathbb E_{\mathbf{X}}\mathbf{X}\mathbf{W}\mathbf{X}^T+tN \mathbb EK_1X_1^+X_1^{+T}\right)^{-1}\mathbb E_{\mathbf{X}}\mathbf{X}\mathbf{W}(\mathbf{Y}_F-\mathbf{Y})\\
            &=\left(1+\frac{tN}n\right) 
            \left(n\mathbb EK_1X_1^+ X_1^{+T}+tN \mathbb EK_1X_1^+X_1^{+T}\right)^{-1}n\mathbb E K_1X_1(F(X_1)-Y_1)\\
            &=\left(\mathbb EK_1X_1^+X_1^{+T}\right)^{-1}\mathbb E K_1X_1(F(X_1)-Y_1).
        \end{split}
    \end{equation*}
    The third and fifth equation hold because the inverse operation of non-singular matrix is continuous. 

    Through the proof of Theorem \ref{thm:estimation} in Section \ref{appsubsec:thm_estimation}, we have $\mathbb E\widehat{\Delta}_{(n)} = \mathbb E\Delta=\mathbb E\widehat{\Delta}^{\text{HD}}$. Thus, we have the expectation of high dimensional form $\mathbb E \widehat{\theta}^{\text{HD}}(t) = \mathbb E\{(\widetilde{\mathbf{X}}\widetilde{\mathbf{W}}\widetilde{\mathbf{X}}^T)^{-1}\widetilde{\mathbf{X}}\widetilde{\mathbf{W}}\widetilde{\mathbf{Y}}_F - \widehat{\Delta}^{\text{HD}} (t)\}=\mathbb E\widehat\theta^{\text{PP}}$.

    About the normality of $\widehat{\theta}^{\text{HD}}$, 

    \begin{equation*}
        \begin{split}
            \widehat{\theta}^{\text{HD}}(t) =& \left(\widetilde{\mathbf{X}}\widetilde{\mathbf{W}}\widetilde{\mathbf{X}}^T\right)^{-1}\widetilde{\mathbf{X}}\widetilde{\mathbf{W}}\widetilde{\mathbf{Y}}_F - \left(\mathbf{X}\mathbf{W}\mathbf{X}^T+t\widetilde{\mathbf{X}}\widetilde{\mathbf{W}}\widetilde{\mathbf{X}}^T\right)^{-1}{\mathbf{X}}{\mathbf{W}}(\mathbf{Y}_F-\mathbf{Y})\\
            =&\left(\sum_{i=1}^N \widetilde K_i\widetilde X_i^+\widetilde X_i^{+T}\right)^{-1}\sum_{i=1}^N \widetilde K_i\widetilde X_i^+F(\widetilde X_i)\\
            &-\left(1+\frac{tN}n\right)\left(\sum_{i=1}^n K_i X_i^+ X_i^{+T}+t\sum_{i=1}^N \widetilde K_i\widetilde X_i^+\widetilde X_i^{+T}\right)^{-1}\sum_{i=1}^n K_i X_i^+(F(X_i)-Y_i).
        \end{split}
    \end{equation*}

    By the normality derived by the central limit theorem, we have 
    \begin{gather*}
        \sum_{i=1}^N \widetilde K_i\widetilde X_i^+F(\widetilde X_i) \to_d N\left(\mathbb EK_1X_1F(X_1), N^{-1}\text{Cov}(K_1X_1F(X_1)\right),\\
        \sum_{i=1}^n K_i X_i^+(F(X_i)-Y_i) \to_d N\left(\mathbb EK_1X_1(F(X_1)-Y_1), n^{-1}\text{Cov}(K_1X_1(F(X_1)-Y_1)\right),\\
    \end{gather*}
    and 
    \begin{gather*}
    \sum_{i=1}^N \widetilde K_i\widetilde X_i^+\widetilde X_1^{+T} \to_p\   S_n= A(h) + O_p\left(\{nh^p\}^{-1/2}\right),\\
    \left(1+\frac{tN}n\right)\left(\sum_{i=1}^n K_i X_i^+X_i^{+T}+t\sum_{i=1}^N \widetilde K_i\widetilde X_i^+\widetilde X_1^{+T}\right) \to_p\  S_n = A(h) + O_p\left(\{nh^p\}^{-1/2}\right).\\
    \end{gather*}
    
    Thus, the multiple of a convergence to constant in probability and a convergence to a Gaussian normality in distribution also converges to a Gaussian distribution, namely
    $$
    \widehat\theta^{\text{HD}}(t) \to_d N\left(\mathbb E\widehat\theta^{\text{HD}}(t), \text{Cov}\left(\widehat\theta^{\text{HD}}(t)\right)\right)=N\left(\theta^*, \text{Cov}\left(\widehat\theta^{\text{HD}}(t)\right)\right).
    $$
    This shows the exactly the same properties with $\widehat{\theta}^{\text{con}}$ and $\widehat{\theta}^{\text{PP}}$.
\end{proof}

\end{document}